\documentclass[runningheads]{llncs}

\usepackage[mobile]{eccv}

\usepackage{eccvabbrv}

\usepackage{graphicx}
\usepackage{placeins}
\usepackage{multirow}
\usepackage{booktabs}
\usepackage{siunitx}
\usepackage{svg} %
\usepackage[nolist,nohyperlinks,printonlyused]{acronym}
\usepackage[symbols,nogroupskip,sort=none]{glossaries-extra} %
\usepackage{soul} %
\newcommand{\pointcloud}{\ensuremath{\mathcal P}}
\newcommand{\PC}{\pointcloud}

\glsxtrnewsymbol[description={}]{fooo}{\bf{Unsupervised Flow Parameters}}

\glsxtrnewsymbol[description={Point Cloud a}]{cloud_a}{\ensuremath{P_a}}
\glsxtrnewsymbol[description={Point Cloud b}]{cloud_b}{\ensuremath{P_a}}
\glsxtrnewsymbol[description={Flow from Point Cloud a to b}]{flow_a_b}{\ensuremath{F_{a\rightarrow b}}}
\glsxtrnewsymbol[description={Flow from Point Cloud b to a}]{flow_b_a}{\ensuremath{F_{b\rightarrow a}}}

\glsxtrnewsymbol[description={ego frame}]{Ei}{\ensuremath{E^i}}
\glsxtrnewsymbol[description={world frame}]{W}{\ensuremath{W}}
\glsxtrnewsymbol[description={point coordinates}]{ps}{\ensuremath{p_S}}

\glsxtrnewsymbol[description={probability that a point disappears going from point cloud A to B}]{disAB}{\ensuremath{\mathcal{D}_{a\rightarrow b}^{(b)}}}
\glsxtrnewsymbol[description={probability that a point disappears going from point cloud B to A}]{disBA}{\ensuremath{\mathcal{D}_{b\rightarrow a}^{(a)}}}

\glsxtrnewsymbol[description={probability that a point is dynamic going from point cloud A to B}]{dynAB}{\ensuremath{\mathcal{M}_{a\rightarrow b}^{(b)}}}
\glsxtrnewsymbol[description={probability that a point is dynamic going from point cloud B to A}]{dynBA}{\ensuremath{\mathcal{M}_{b\rightarrow a}^{(a)}}}

\glsxtrnewsymbol[description={probability that a point is static going from point cloud A to B}]{statAB}{\ensuremath{\mathcal{S}_{a\rightarrow b}^{(b)}}}
\glsxtrnewsymbol[description={probability that a point is static going from point cloud B to A}]{statBA}{\ensuremath{\mathcal{S}_{b\rightarrow a}^{(a)}}}

\glsxtrnewsymbol[description={probability that a point is a ground point going from point cloud A to B}]{grndAB}{\ensuremath{\mathcal{G}_{a\rightarrow b}^{(b)}}}
\glsxtrnewsymbol[description={probability that a point is a ground going from point cloud B to A}]{grndBA}{\ensuremath{\mathcal{G}_{b\rightarrow a}^{(a)}}}

\glsxtrnewsymbol[description={complete Pose, where S denotes the coordinate system in which we measure the coordinates \gls{ps}}]{PS}{\ensuremath{P_S}}

\glsxtrnewsymbol[description={ego position}]{e}{\ensuremath{e}}
\glsxtrnewsymbol[description={time}]{ti}{\ensuremath{t_i}}
\begin{acronym}

\acro{kittisf}[KITTI-SF]{KITTI Stereo Flow}

\acro{bev}[BEV]{Bird's-Eye-View}
\acro{fov}[FoV]{field of view}
\acro{cnn}[CNN]{convolutional neural network}
\acro{dl}[DL]{deep learning}
\acro{pfe}[PFE]{point feature encoder}
\acro{pfn}[PFN]{Pillar Feature Net}
\acro{tcr}[TCR]{Track Check Repeat}
\acro{sct}[SCT]{Sample Crop Track}
\acro{gru}[GRU]{Gated Recurrent Unit}
\acro{flop}[FLOP]{floating-point operation}
\acro{nds}[NDS]{NuScenes detection score}
\acro{map}[mAP]{mean average precision}
\acro{pp}[PP]{PointPillars}
\acro{ad}[AD]{autonomous driving}
\acro{av}[AV]{autonomous vehicle}
\acro{sota}[SOTA]{state-of-the-art}

\acro{DNN}[DNN]{deep neural network}

\acro{MLP}[MLP]{Multi-layer perceptron}

\end{acronym}

\newcommand{\avv}{AV2}
\newcommand{\ad}{\ac{ad}}

\newcommand{\sota}{\ac{sota}}

\newcommand{\bev}{\ac{bev}}

\newcommand{\fov}{\ac{fov}}

\newcommand{\kitti}{KITTI}

\newcommand{\nusc}{nuScenes}
\newcommand{\wod}{WOD}

\newcommand{\ssup}{\text{self-supervised}}
\newcommand{\lidar}{\text{lidar}}
\newcommand{\lsf}{lidar scene flow}
\newcommand{\pgt}{pseudo ground truth}

\newcommand{\dbscan}{\text{DBSCAN}}

\newcommand{\bff}{\mathbf{f}} %

\newcommand{\bI}{\mathbf{I}}

\newcommand{\bp}{\mathbf{p}}

\newcommand{\bT}{\mathbf{T}}

\newcommand{\bx}{\mathbf{x}}

\newcommand{\nR}{\mathbb{R}}

\newcommand{\cB}{\mathcal{B}}

\newcommand{\cX}{\mathcal{X}}

\newcommand{\figref}[1]{Fig.~\ref{#1}}
\newcommand{\secref}[1]{Section~\ref{#1}}

\newcommand{\tabref}[1]{Table~\ref{#1}}

\def\wrt{w.r.t.}

\newcommand{\boldparagraph}[1]{\vspace{0.2cm}\noindent{\bf #1:} }

\definecolor{darkyellow}{rgb}{1.0, 0.75, 0.0}
\definecolor{darkgreen}{rgb}{0,0.7,0}

\newcommand{\red}[1]{\noindent{\color{red}{#1}}}

\newcommand{\yellow}[1]{\noindent{\color{darkyellow}{#1}}}

\soulregister\cite7

\ifdefined\DIFFVERSION

\fi

\ifdefined\OLDVERSION

\fi

\ifdefined\NEWVERSION

\fi

\newcommand{\R}[1]{{%
    \textbf{%
        \ifstrequal{#1}{1}{\textcolor{red}{R#1}}{%
        \ifstrequal{#1}{2}{\textcolor{blue}{R#1}}{%
        \ifstrequal{#1}{3}{\textcolor{magenta}{R#1}}{%
        \ifstrequal{#1}{4}{\textcolor{teal}{R#1}}{%
                           \textcolor{cyan}{R#1}%
        }}}}%
    }%
}}

\usepackage[accsupp]{axessibility}  %

 \usepackage[pagebackref,breaklinks,colorlinks,citecolor=eccvblue]{hyperref}

\usepackage{orcidlink}

\begin{document}

\title{LISO: Lidar-only Self-Supervised 3D Object Detection}

\author{Stefan Andreas Baur\inst{1,2} \and
Frank Moosmann\inst{1} \and
Andreas Geiger\inst{2,3}
}

\authorrunning{S.~Baur et al.}

\institute{Mercedes-Benz, Germany\and
University of Tübingen\and
Tübingen AI Center}

\maketitle
\begin{abstract}
3D object detection is one of the most important components in any Self-Driving stack, but current \sota\ \lidar~object detectors require costly \& slow manual annotation of 3D bounding boxes to perform well.
Recently, several methods emerged to generate \pgt~without human supervision, however, all of these methods have various drawbacks:
Some methods require sensor rigs with full camera coverage and accurate calibration, partly supplemented by an auxiliary optical flow engine.
Others require expensive high-precision localization to find objects that disappeared over multiple drives.

We introduce a novel self-supervised method to train \sota~\lidar~object detection networks which works on unlabeled sequences of lidar point clouds only, which we call trajectory-regularized self-training.
It utilizes a \sota~self-supervised \lsf~network under the hood to generate, track, and iteratively refine \pgt.
We demonstrate the effectiveness of our approach for multiple \sota~object detection networks across multiple real-world datasets.
Code will be released.

\keywords{Self-Supervised \and LiDAR \and Object Detection}
\end{abstract}

\section{Introduction}
\label{sec:intro}

\begin{figure}
\begin{center}
\resizebox{.99\textwidth}{!}{%
\includegraphics[height=5cm, trim=3cm 12cm 8cm 10cm,clip]{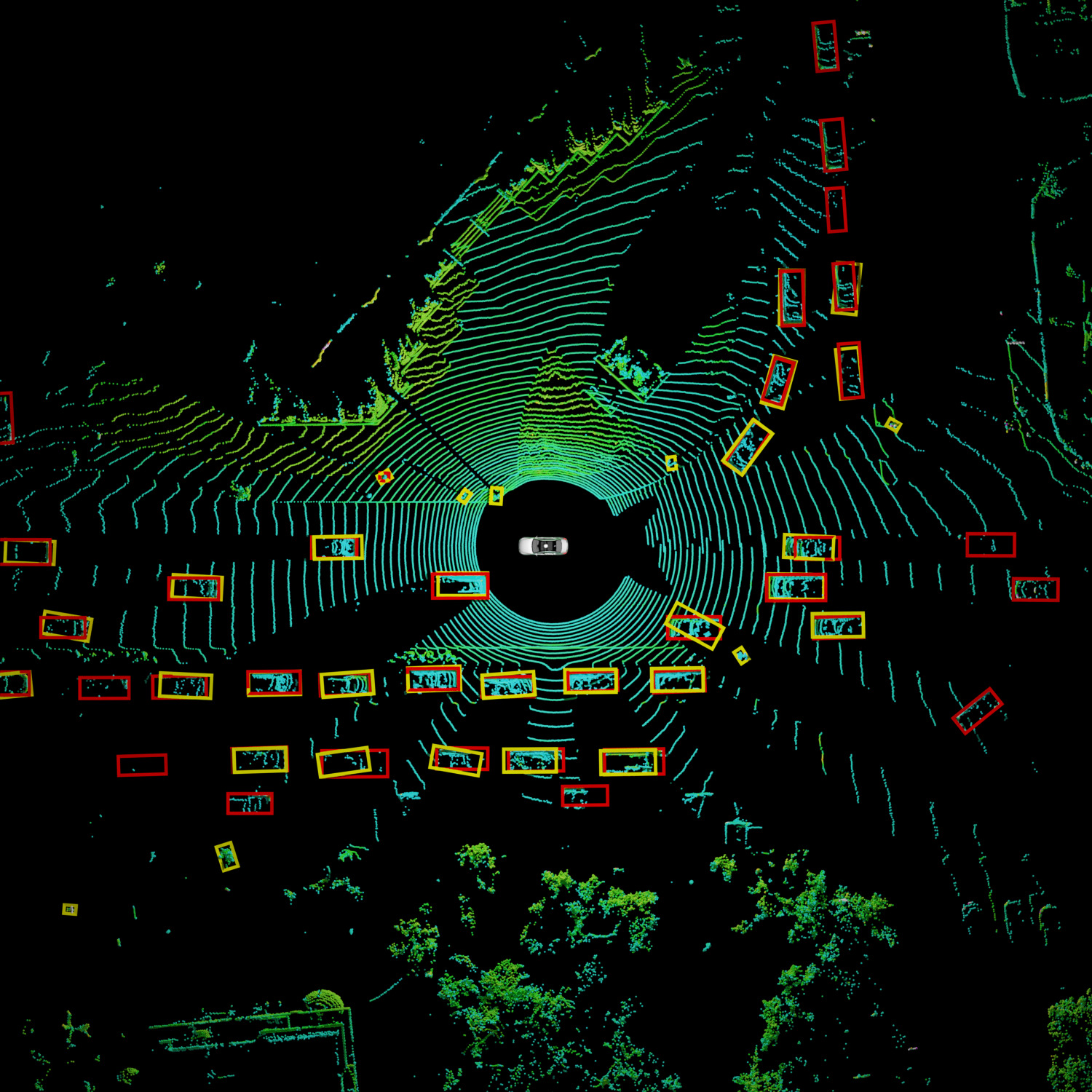}%
\hspace{-0.5mm}
\includegraphics[height=5cm]{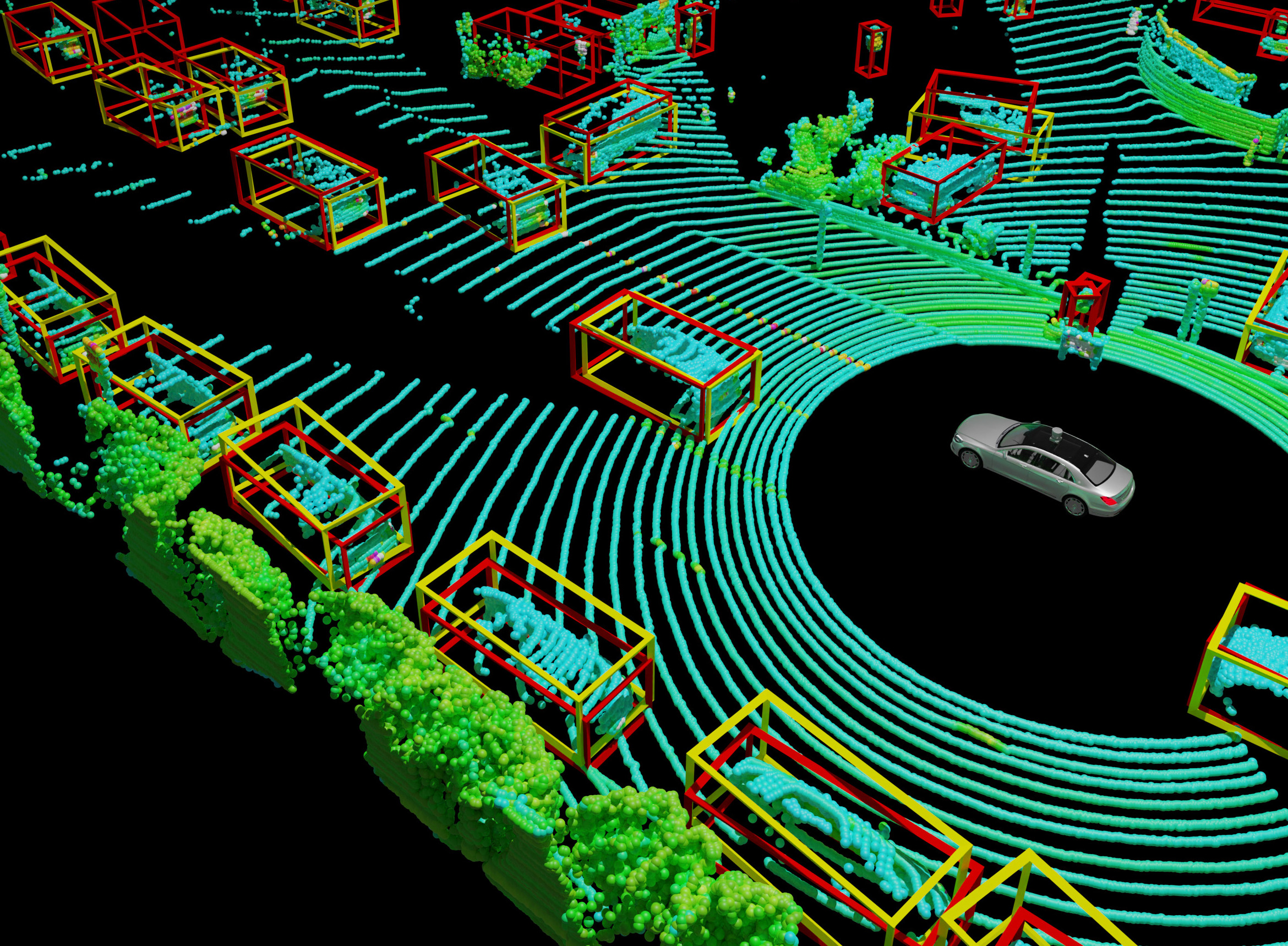}%
}
\caption{\textbf{Objects predicted by our method using no manual annotations.}
\red{Red} boxes are ground truth boxes, \yellow{yellow} boxes are predicted by our network. }
\label{fig:eyecatcher}
\end{center}
\vspace{-3mm}
\end{figure}

Human developmental research reveals that infants less than one year of age are able to categorize animate and inanimate objects based on observed motion cues and can generalize this categorization to previously unseen objects \cite{speed_and_direction_animacy}.
Yet, lidar object detectors with \sota\ performance are trained using manually selected and categorized annotations.
These annotations are very expensive to obtain and become outdated quickly: new lidar sensors are coming to market regularly, trained \sota\ object detectors are sensitive to sensor characteristics and change in mounting position, and existing annotations are difficult to transfer between sensors or different sensor mounting positions \cite{rist}, resulting in high re-labeling efforts with each change.

In this paper, we aim at bridging this gap by distilling motion cues observed in \ssup\ \lsf\ into \sota\ single-frame \lidar\ object detectors. 
We introduce a novel \ssup\ method which is charmingly simple and easy to use:
Input to our method are solely lidar pointcloud sequences.
No human annotated bounding boxes, cameras, costly high-precision GPS, or tedious sensor rig calibration is required.
Our method trains and runs a \ssup\ lidar flow estimator~\cite{slim} under the hood in order to create motion cues.
We show in our experiments that these motion cues are a key factor to the success of our method.
Based on the estimated lidar flow we bootstrap initial \pgt\ using simple clustering and track optimization. 
With these mined bounding boxes of \textit{moving} objects we initialize a self-supervised, trajectory-regularized variant of self-training~\cite{selftraining} (which is semi-supervised in its original form):
We train a first version of a \sota\ object detector, then iteratively re-generate and trajectory-regularize the \pgt\ and re-train the detector.
Since the single-frame object detector has no concept of motion, it generalizes to detect any \textit{movable} object on the way.
Exemplary output of the trained detector, 3D boxes in single-frame point clouds, is depicted in \figref{fig:eyecatcher}.

Our contribution is a novel \ssup\, trajectory-regularized self-training framework for single-frame 3D object detection with the following properties:
\begin{itemize}
    \item It is based entirely on \lidar, i.e. without the limitations of prior works: no cameras, no calibration, no high-precision GPS, no manual annotations.
    \item It is agnostic \wrt\ the sensor model, mounting pose, and detectors architecture and works with the same set of hyper parameters: We demonstrate this across four different datasets and different \sota\ detector networks.
    \item It is able to generalize from \textit{moving} objects (motion cues) to \textit{movable} objects (final detection results) and significantly outperforms \sota\ methods while being simpler.
\end{itemize}
We show that using motion cues together with trajectory-regularized self-training is key to this success.

The code of this approach will be published for easy use and comparison by other researchers.

The paper is organized as follows:
Sec.~\ref{sec:related} discusses related work before the proposed method is described in detail in Sec.~\ref{sec:method}. Extensive experiments in Sec.~\ref{sec:eval} show the performance of our method before we conclude in Sec.~\ref{sec:conclusion}.

\section{Related Work}
\label{sec:related}

\subsection{Single Frame Lidar 3D Object Detection}
Object detectors operating on 3D point clouds are an active research field.
The currently best-performing ones are using deep neural networks trained via supervised learning and can be categorized by their internal representation:
Some networks operate on points directly like PointRCNN \cite{pointrcnn}, 3DSSD \cite{3dssd}, and IA-SSD \cite{iassd}.
Others project the points either to a virtual range image \cite{lasernet,rsn,rangeioudet} or into a voxel representation like VoxelNet \cite{voxelnet}, PointPillars, CenterPoint \cite{centerpoint}, and Transfusion-L \cite{transfusion}. 
However, all aforementioned methods require large human-annotated datasets in order to perform well and obtaining such annotations is very expensive.
We address this problem in this paper, enabling training of \sota~object detectors using \pgt.

\subsection{Object Distilling from Motion Cues} %
\label{sec:distilling}
Multiple approaches have been suggested which leverage motion cues from lidar frame pairs in order to detect \textit{moving} objects.
Using the assumption of local geometric constancy, they decompose dynamic scenes into separate moving entities by applying as-rigid-as-possible optimization. Examples of these are the works by Dewan et al.~\cite{dewan}, RSF~\cite{rsf} and OGC~\cite{ogc}. The two former methods are optimization-based whereas OGC uses these constraints as loss-function to self-supervise a segmentation network.

Although these methods avoid the need for expensive labels, in contrast to our method they tend not to work well in low-resolution areas, they are typically slow, and can only detect \textit{moving} but not \textit{movable} (\ie static but potentially moving) objects.

\subsection{Pseudo Ground Truth for Object Detection}
Different approaches have been proposed to mine \pgt\ for training object detectors:
Najibi et al.~\cite{waymo} and very recently Seidenschwarz et al.~\cite{semoli} use motion cues similar to section~\ref{sec:distilling} in order to distill \textit{moving} objects as \pgt\ and use it to train an object detector. \cite{waymo} runs optimization for each frame pair to obtain \lsf, clusters points, fits boxes, tracks them using a Kalman Filter and finally refines them on ICP-registered point clouds. \cite{semoli} optimizes a clustering algorithm on motion cues through message passing which they then apply to segment point clouds into a set of instances and subsequently fit boxes. As shown in \cite{semoli}, and in contrast to our approach, both methods suffer from a large performance gap between \textit{moving} and \textit{movable} objects.
We demonstrate that our method does not suffer from this gap.

\cite{tcr}, \cite{sct} were the first to \textit{iteratively} apply a detection, tracking, retraining paradigm to \ad~data.
They build upon consistency constraints between object detectors which they train in the lidar and camera domain, making use of an optical flow network which is trained using supervision.
The approach requires a calibrated sensor rig with possibly large camera coverage, IMU, and precision GPS.
Similarly, \cite{WangCZ22} uses video sequences together with \lsf\ to jointly train a camera and a lidar object detector.
MODEST \cite{modest,hypermodest} does not require coverage by calibrated cameras, but instead
adds the additional requirement to have multiple lidar recordings of the same location in order to identify objects that vanished over time.
With such demanding requirements, these approaches are not easy to use and are partially unsuited for popular \ad~datasets.

Oyster~\cite{oyster} is the approach most similar to ours.
It uses \dbscan~\cite{dbscan} on \lidar\ point clouds to initialize \pgt.
Clusters are then tracked using forward-and-reverse tracking in sensor coordinates (i.e. without consideration for ego motion), using a complex policy for confidence-based track retention.
After training an object detector on close range data, they employ zero-shot generalization to the far range data, track again and iteratively retrain the detector.
In their experiments they use different hyperparameters for different datasets.
Our method, in contrast, explicitly considers sensor motion, produces much cleaner initial proposals by leveraging a \ssup\ \lsf~network, does not require zero-shot-generalization from near-to-far-field, and works with current \sota~object detectors. We show on multiple datasets that our method outperforms \cite{oyster}, that it works robustly with the same set of hyper-parameters and also generalizes well to detect \textit{movable} objects.
Additionally,
we will release our code to facilitate further research on this topic.

\section{Method}
\label{sec:method}
\begin{figure*}[ht]
\centering
\includegraphics[width=\textwidth]{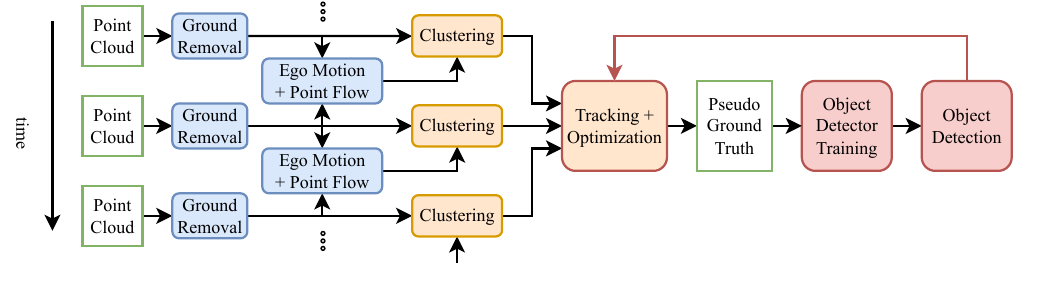}
  \vspace{-8mm}
  \caption{\textbf{Overview of the proposed method.} Point cloud sequences are preprocessed (blue, Sec~\ref{sec:preprocessing}), initial \pgt\ is created (orange, Sec.~\ref{sec:pgtgeneration}) and the object detector is iteratively trained and \pgt\ regenerated (red, Sec.~\ref{sec:pgtgeneration}).}
  \vspace{-3mm}
\label{fig:overview}
\end{figure*}

\begin{figure*}[ht]
    \centering
    \includegraphics[width=0.49\textwidth]{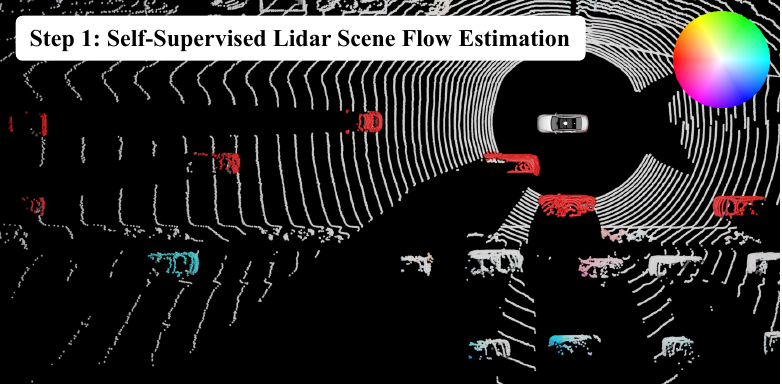}
    \includegraphics[width=0.49\textwidth]{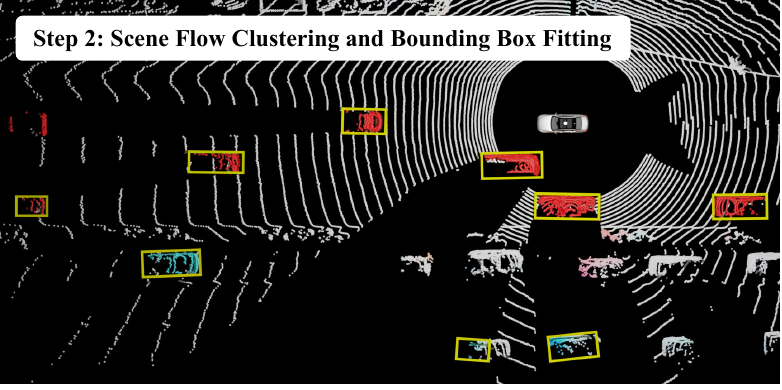}\\
    \includegraphics[width=0.49\textwidth]{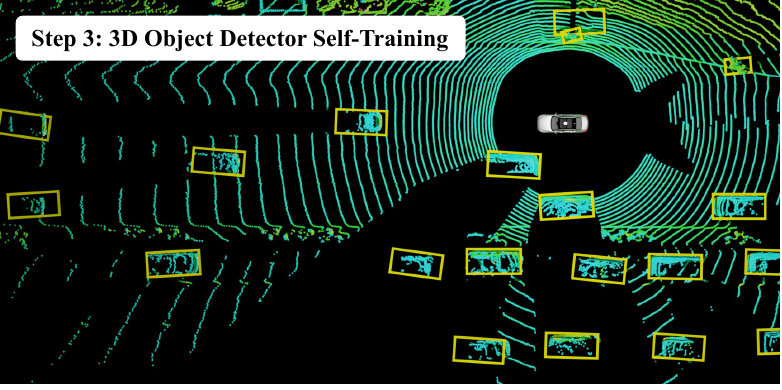}
    \includegraphics[width=0.49\textwidth]{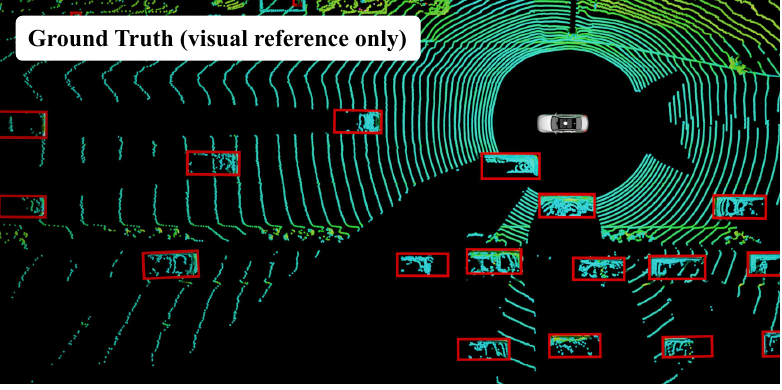}
    \caption{
    \textbf{Overview over preprocessing, initial \pgt\ generation and training with examples.}
    \textbf{Top left:} In the first step, \ssup\ \lsf\ is computed and corrected for vehicle ego-motion. Points are colored by flow direction and magnitude.
    \textbf{Top right:} In the second step, the scene flow is clustered and bounding boxes are fitted (to the moving objects). 
    \textbf{Bottom left:} In the third step, the network is trained on the \pgt\ and is generalizing to static objects also, since it does not have the motion information as input signal.
    Points are thus colored by laser intensity.
    \textbf{Bottom right:} Ground truth, for reference.
    }
    \vspace{-3mm}
    \label{fig:processingsimages}
\end{figure*}

A general overview of our method is sketched in 
Fig.~\ref{fig:overview} and some steps illustrated in Fig.~\ref{fig:processingsimages}. As input, we take raw (unlabeled) point cloud sequences and undergo three stages, all of which are detailed in the following: Preprocessing of point clouds and \lsf\ computation (Sec.~\ref{sec:preprocessing}), initial \pgt\ generation (Sec.~\ref{sec:pgtgeneration}) and repeated training with \pgt\ refinement (Sec.~\ref{sec:iterativetraining}).
The final output of the method is a trained object detector which can detect \textit{movable} objects in raw single-frame point clouds.

\subsection{Preprocessing}
\label{sec:preprocessing}
We preprocess the raw input point clouds as follows:

\boldparagraph{Ground Removal}
First, we remove distracting ground points from each single point cloud using JCP \cite{groundseg}, which is a simple, robust, yet effective algorithm to remove ground points using changes in observed height above ground.

\boldparagraph{Ego Motion Estimation}
Second, we compute ego-motion between neighboring frame pairs using KISS-ICP~\cite{kiss}, which is based on a robust version of ICP.
The output is a cm-level accurate transformation \(\bT_{\text{ego}}^{t\rightarrow t+1} \in \nR^{4\times4}\), describing the ego vehicle position at time \(t+1\) represented in the ego frame from time \(t\).

\boldparagraph{Lidar Scene Flow Estimation}
Third, we compute \lsf\ between neighboring frame pairs resulting in a flow vector \(\bff_i = (dx, dy, dz)\) for every point $i$ in the first point cloud \(\PC^{t}\).
We chose to use SLIM~\cite{slim} as its code is readily available, it is easy to use, features fast inference, and produces \sota\ results.
The network is trained \ssup\ on raw point cloud sequences, minimizing a k-nearest-neighbor loss between forward and time-reversed point clouds.

The components of our preprocessing steps (Ground Removal, Ego Motion Estimation, Lidar Scene Flow Estimation) have been selected for robustness and are all used with their default parameters from their respective publications.

\subsection{Initial Pseudo Ground Truth Generation}
\label{sec:pgtgeneration}
The aim of our method is to mine \pgt\ for training a 3D lidar object detector. For a single point cloud \(\PC^{t}\) (consisting of $n\in\mathbb{N}$ points $\bp_i$, $\bp_i\in\PC^{t}$) this ground truth is a set of 3D bounding boxes $\cB$ with confidences representing the objects at time $t$:
\begin{equation}
    \cB^{t} =\{\text{B}^{t}_{j}, j \in \mathbb{N}\} = \{(x, y, z, l, w, h, \theta, c)_j\}
\end{equation}
Here, \((x, y, z)=\bx\) define the center position, \((l, w, h)\) the length, width and height, \(\theta\) the heading (orientation around up axis), and \(c\) the confidence for a single box.

The key success factor of our method is to focus on a high precision (and potentially low recall) of the initial set of bounding boxes in order to avoid "wrong" objects to negatively influence the object detector.
We achieve this by leveraging shape, density, and especially motion cues to robustly identify \textit{moving} objects solely (see also top-right of Fig.~\ref{fig:processingsimages}).
Sec.~\ref{sec:iterativetraining} targets at generalizing these to \textit{movable} objects later on.

\boldparagraph{Flow Clustering}
The points $\bp_i\in\PC^{t}$ in each preprocessed point cloud are clustered based on geometry and motion:
All stationary points in a scene should have a flow \(\bff_{i}\) similar to the vehicle's ego-motion, i.e. \(\bff_{i} \approx \bff_{i,\text{sta}} = ((\bT_{\text{ego}}^{t\rightarrow t+1})^{-1} - \bI_4)\cdot\bp_i\).
The residual flow must then be caused by motion of other actors: \(\bff_{i,\text{dyn}} = \bff_{i} - \bff_{i,\text{sta}} \).
We filter all static points by applying a threshold of 1\si{\m/\s} to the residual flow and cluster the remaining points based on their point location and flow vector in 6D using \dbscan~\cite{dbscan}(with parameters $\varepsilon=1.0$, $\text{minPts}=5$).

We fit a 3D bounding box $\text{B}^{t}_{j}$ to each resulting cluster following \cite{modest} and discard boxes with $l/w>4.0$, $lw<0.35\si{\m}^2$, or $lwh<0.5\si{\m}^3$.
The heading  \(\theta\) is set to match the "forward-axis" of the motion, i.e. we orient the boxes along the direction of the residual flow \(\bff_{i,\text{dyn}}\).
The confidence \(c\) is set to 1.

\boldparagraph{Tracking}
We run a simple flow based tracker: Since we have accurate ego-motion available, we can track accurately in a fixed coordinate system \wrt~the world. 
First, using the residual flow, we can compute for each box at $t$ a rigid body transform that transforms the proposed box $\text{B}_i^t$ forward in time towards its suspected location at $t+1$, $\hat{\text{B}}_i^{t+1}$.
The propagated boxes $\hat{\cB}^{t+1}$ are matched greedily against the actual boxes $\cB^{t+1}$ based on box-center distance to the new detections.
Unmatched boxes in $\cB^{t+1}$ which are further away than 1.5m from propagated boxes spawn new tracklets.
Unmatched tracklets from $\cB^{t}$ are propagated according to the last observed box motion for up to one time step, after that unmatched tracklets are terminated.
We run tracking forward and reverse in time, connecting tracklets from forward and reverse tracking to tracks.

The resulting set of tracks is post-filtered: We discard tracks that are shorter than 4 time steps or that have a median box confidence $c$ lower than threshold $0.3$ (note that the initial confidence $c=1$ set during clustering is later replaced by detectors confidences, see Sec.~\ref{sec:iterativetraining}). This retains only stable and high-confident tracks and avoids false positives to enter the \pgt.

\boldparagraph{Track Optimization}
We reduce positional noise of the tracks by minimizing translational jerk on all tracks longer than 3\si{\m}.
Let $\cX_{\text{obs}}$ be the sequence of (noisy) observed box center positions $\bx_i$ for consecutive time steps $i\in \{1,...,T\} $ of a track and their derivative $\frac{d\bx_i}{dt} \approx \frac{\bx_{i+1}-\bx_{i}}{\Delta t}$.
We compute smoothed track positions $\cX_{\text{smooth}}$ by initializing them to $\cX_{\text{obs}}$ and minimizing the following loss w.r.t. $\bx_{\text{smooth}}$:
\begin{equation}
L = \sum_{i=1}^{T} \left\Vert  \frac{d^4\bx_{i,\text{smooth}}}{dt^4} \right\Vert_2 ^2 + \beta  \left\Vert \bx_{i,\text{smooth}} - \bx_{i,\text{obs}}\right\Vert_2 ^2
\end{equation}
I.e. we minimize the jerk $\frac{d^4\bx}{dt^4}$ and use a quadratic regularizer term ($\beta=3$) on the positions. Experiments revealed that this simple optimization goal outperforms more sophisticated ones like fitting an unrolled bicycle model or adding terms for acceleration, which leads to tracks "overshooting" corners, especially when using aggressive optimization parameters required to run at reasonably fast optimization time.

We subsequently align the orientation $\theta$ of each detection in a track to the direction of the smoothed track at its position.
Box dimensions $\{l, w, h\}$ of all boxes in a track are adapted to the 90 percentile of observed box dimensions in a track.
All hyperparameters related to clustering, tracking and track optimization have been tuned visually on two sample snippets from \nusc.

\subsection{Trajectory-Regularized Self-Training}
\label{sec:iterativetraining}
After having mined initial \pgt\ of \textit{moving} objects with a high precision and low recall, we now aim at iteratively improving our \pgt\ by training and using an object detector so that the \pgt\ generalizes to \textit{movable} objects.
We achieve this by executing iterative trajectory-regularized self-training, which is composed of the following two steps:

\boldparagraph{Training}
We train the target object detection network using the current \pgt\ in a supervised training setup. Any single-frame object detection network can be plugged into our pipeline.
In our experiments we do not deviate from the basic training setup of our object detection networks.
Like any SOTA object detection method, we apply standard augmentation techniques to a point cloud during network training: Random rotation, scaling ($\pm5\%$), and random translation up to $5\si{\m}$ around the origin.
Furthermore, we randomly pick 1 to 15 objects from the \pgt~database and insert a random subset of their points at random locations into the scene.

\boldparagraph{Pseudo Ground Truth Regeneration}
After a certain amount of training steps ($s=30k$ in the experiments) we stop the training and use the trained object detector to recreate new, improved \pgt:
We run the trained detector in inference mode over all sequences in the training dataset to generate new box detections $\cB$, thereby using the detectors' confidence as box confidence $c$.
We regularize these detections by running the flow-based tracker and track smoothing exactly like we do for the initial \pgt\ generation.
Every $2nd$ iteration (i.e. every $60k$ steps in the experiments), we discard the network weights after regenerating the \pgt.

\begin{figure}%
    \centering
    \subfloat[\centering Effect of \ssup\ self-learning]{{\includegraphics[width=6cm]{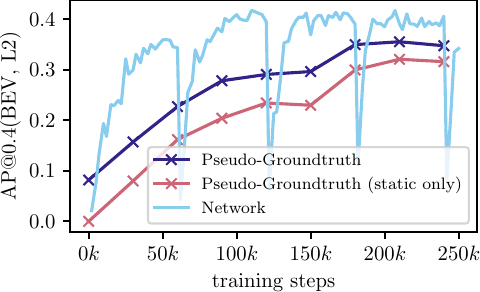} }}%
    \subfloat[\centering Evaluation of Self-Training Hyperparameters]{{\includegraphics[width=6cm]{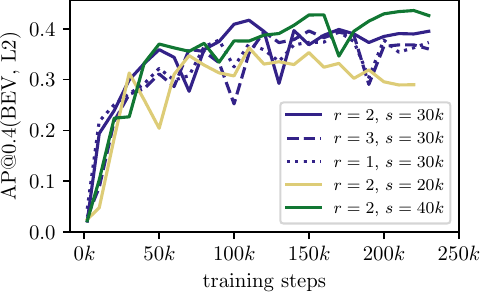} }}%

\caption{
\textbf{Left: Effect of \ssup\ self-learning:} 
At the beginning of the training, only \textit{moving} objects are contained within the \pgt\ and the \pgt\ thus scores 0 on static objects.
Thanks to the self-training, the network generalizes to \textit{movable} objects and the score of the \pgt\ on static objects starts to increase with every regeneration at each ”X”.
The \pgt's performance is measured on a subset of the train set of \wod\ dataset, while the network (here: Centerpoint) is evaluated on a small subset of the validation set.
\textbf{Right: Evaluation of Self-Training Hyperparameters}: Network (Centerpoint) performance over the course of a training on \wod.
$s$ is the number of training steps between \pgt\ regenerations, $r$ is the number of regenerations after which network weights are dropped and re-initialized.
}
\label{fig:pgtNetQuality}
\end{figure}

Fig.~\ref{fig:pgtNetQuality} illustrates the effect of our self-training: We see that the \pgt\ improves with each re-generation and that dropping network-weights allows the network to re-focus on the generalized \pgt. The \pgt\ has a consistently lower performance because we keep it conservative, keeping only highly certain boxes.

Two aspects are key to making our method perform well:
\begin{itemize}
    \item Being restrictive when composing the \pgt\ (i.e. using plausible tracks of high-confidence network detections or clusters with significant flow only) avoids adding false positives into the \pgt\ and hence avoids that our network increasingly hallucinates with each iteration.
    \item Not using flow but only single-frame pointclouds as input for the detector allows the detector to focus on appearance of objects solely.
\end{itemize}
These allow our method to generalize from initially mined \textit{moving} objects to \textit{movable} objects in the scene.

\begin{figure*}[ht!]
    \centering
    \includegraphics[width=0.49\textwidth, trim=0 0 0 24cm,clip]{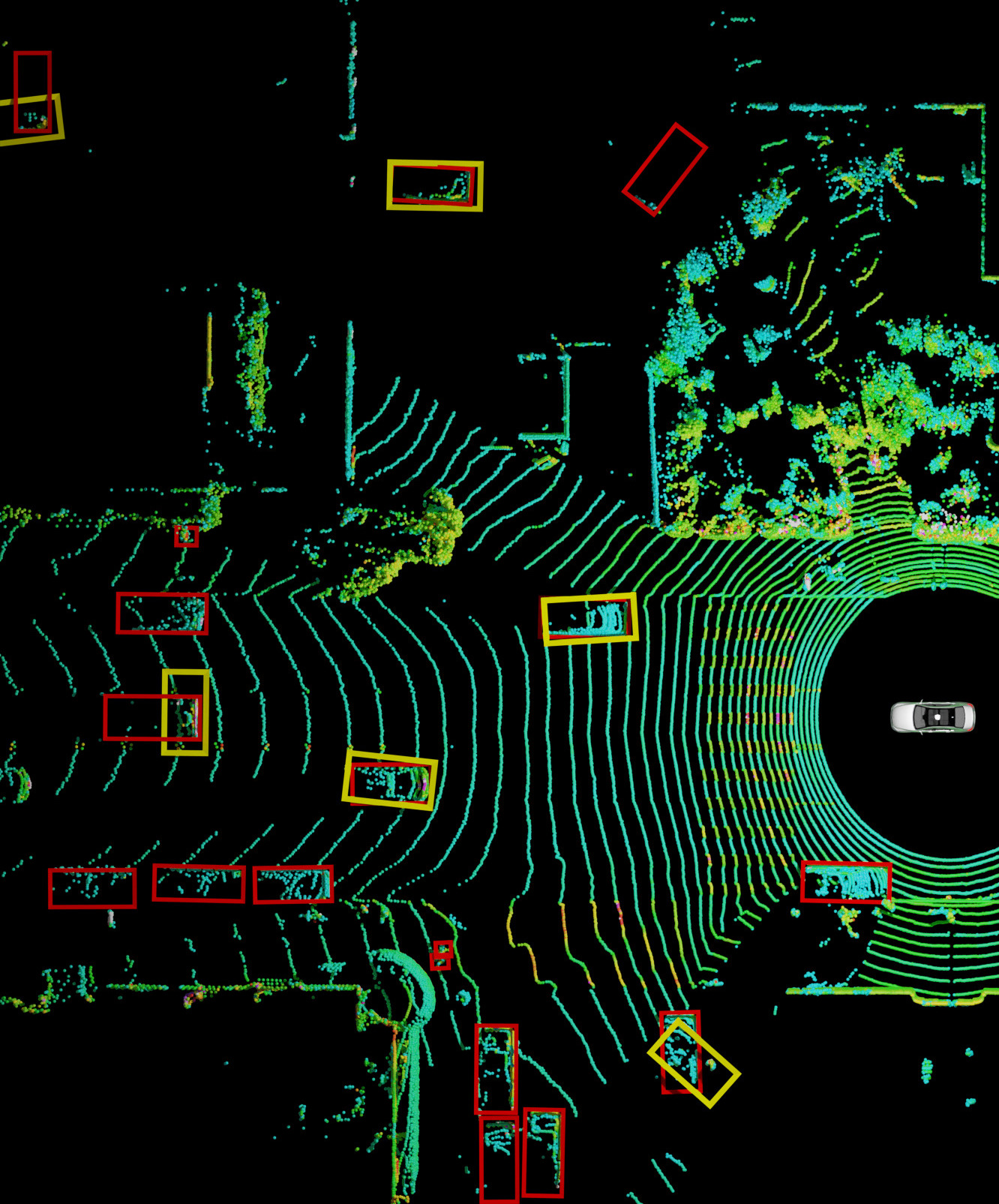}
    \includegraphics[width=0.49\textwidth, trim=0 0 0 24cm,clip]{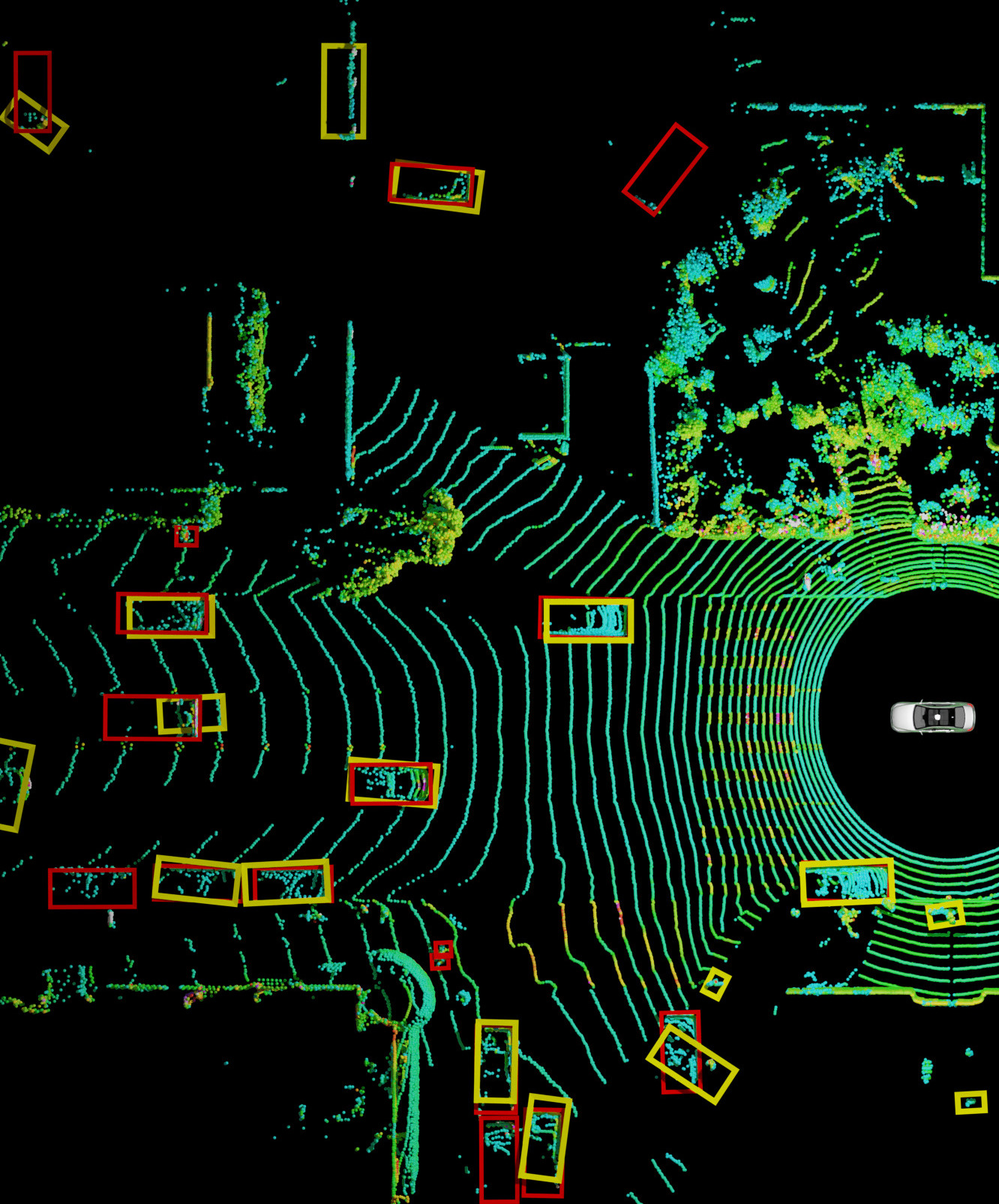}
    \\
    \includegraphics[width=0.49\textwidth, trim=0 4cm 0 0,clip]{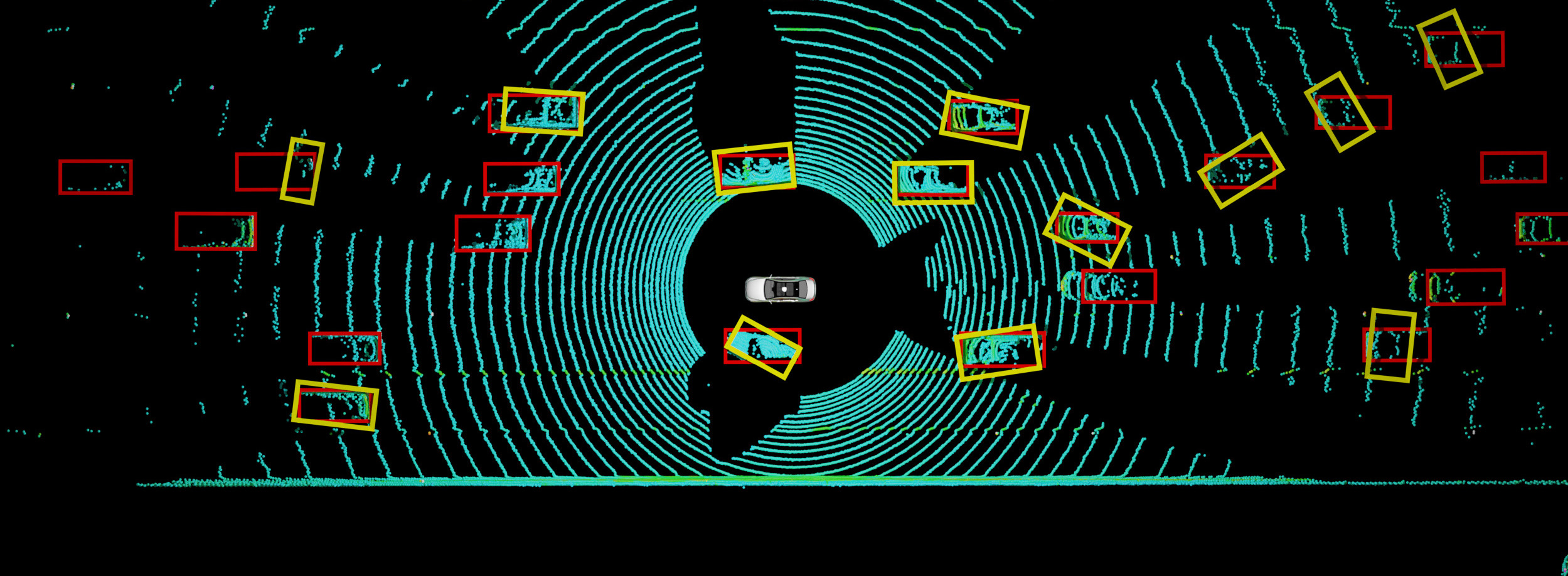}
    \includegraphics[width=0.49\textwidth, trim=0 4cm 0 0,clip]{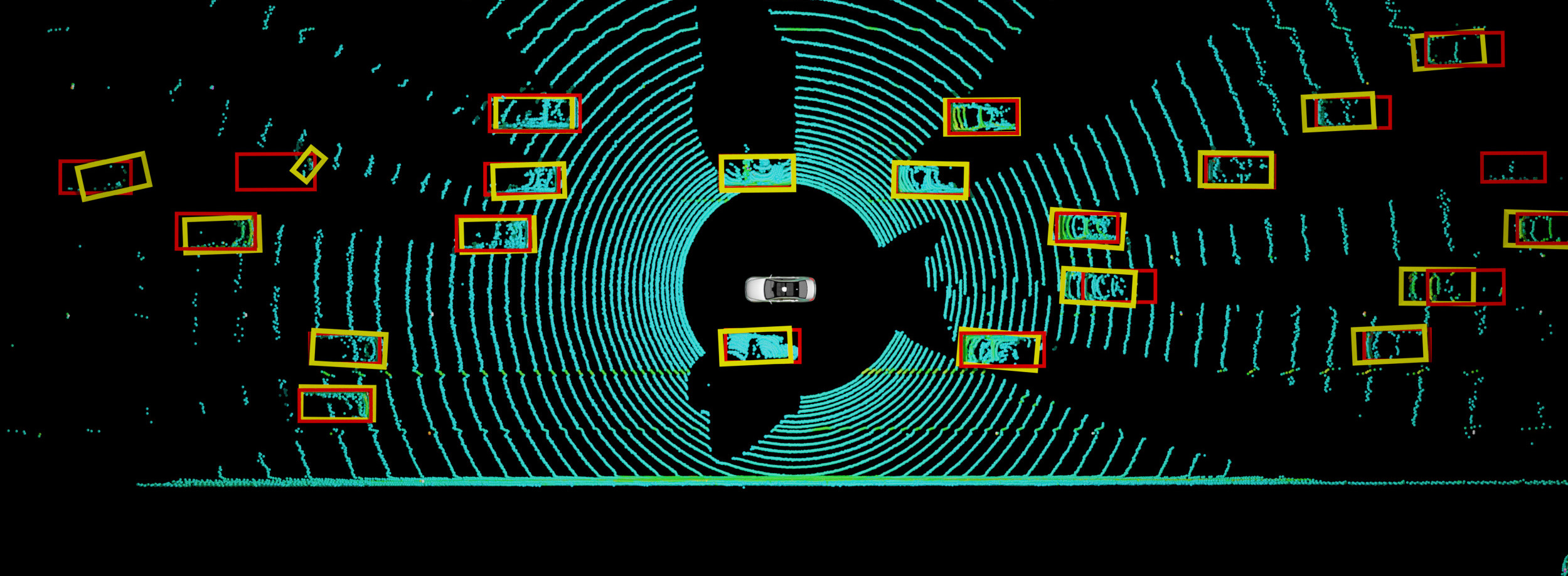}
    
    \caption{
    \textbf{Qualitative Results on \wod.} \red{Red} boxes are ground truth boxes, \yellow{yellow} are predictions. 
    \textbf{Left:} OYSTER-CP
    \textbf{Right:} LISO-CP
    }
    \label{fig:qualitative}
\end{figure*}

\section{Evaluation}
\label{sec:eval}
We evaluate our method on multiple datasets across multiple \sota\ networks.
SLIM \cite{slim}, the \ssup\ flow network we use, is trained and inferred as in the published version, but \bev\ range is extended from $70\times70\si{\m}$, $640\times640$ pixels to $120\times120\si{\m}$ and $920\times920$ pixels.

\begin{table}[t]
\centering
\begin{tabular}{@{}ll|cccc|cc@{}}
\toprule
  \multicolumn{2}{c|}{} &
  \multicolumn{4}{c|}{\textbf{\avv}} &
  \multicolumn{2}{c}{\textbf{\nusc}} \\
  \multirow{2}{*}{} &
   &
  \multicolumn{2}{c}{BEV-IOU} &
  \multicolumn{2}{c|}{3D-IOU} \\
   &
   &
  AP@0.3$\uparrow$ &
  AP@0.5$\uparrow$ &
  AP@0.3$\uparrow$ &
  AP@0.5$\uparrow$ & 
  AP$\uparrow$ & 
  AOE$\downarrow$ \\
\midrule
\parbox[t]{2mm}{\multirow{2}{*}{\rotatebox[origin=c]{90}{\textcolor{gray}{\tiny{Sup.}}}}} 
& CP\cite{centerpoint}  & 0.778 & 0.778 & 0.736 & 0.517 & 0.484 & 0.560  \\
& TF\cite{transfusion}  & 0.783 & 0.654 & 0.767 & 0.601 & 0.627 & 0.501 \\ 

\midrule

\parbox[t]{2mm}{\multirow{2}{*}{\rotatebox[origin=c]{90}{\textcolor{gray}{\tiny{Unsupervised}}}}} 
& \dbscan \cite{dbscan}     & 0.054 & 0.026 & 0.020 & 0.003 & 0.008  & 3.120 \\ 
& \dbscan(SF)  & 0.070 & 0.026 & 0.065 & 0.015 & 0.003  & 2.623 \\
& \dbscan(GF)  & 0.105 & 0.041 & 0.088 & 0.024 & 0.000  & 1.000 \\
& RSF\cite{rsf}          & 0.074 & 0.026 & 0.055 & 0.014 & 0.019  & 1.003  \\
\midrule
\parbox[t]{2mm}{\multirow{5}{*}{\rotatebox[origin=c]{90}{\textcolor{gray}{\tiny{Self Train}}}}} 
& Oyster\cite{oyster} $\dag$    & 0.354 & 0.245 & - & - & - & - \\
& Oyster-CP\cite{oyster}        & 0.381 &  0.266 & 0.150 & 0.002  & 0.091  & 1.514  \\
& Oyster-TF\cite{oyster}        & 0.340 & 0.198 & 0.182 & 0.016  & 0.093  & 1.564 \\
& \textbf{LISO}-CP & \textbf{0.448}   & \textbf{0.335} & \textbf{0.367}    & \textbf{0.188}  & 0.109  & 1.062 \\
& \textbf{LISO}-TF & 0.417 & 0.294 & 0.317 & 0.176 & \textbf{0.134}  & \textbf{0.938} \\

\bottomrule
\end{tabular}%
\caption{\textbf{Evaluation on \avv\ \cite{Argoverse2} and \nusc\cite{caesar2020nuscenes}}:
CP, TF: network architecture, in the first two lines trained supervised for comparison.
$\dag$: Metrics as reported in \cite{oyster}.  SF: self-supervised \lsf\ by SLIM, GF: ground truth \lsf.
Note that \nusc\ uses a minimum precision and recall threshold of 0.1, and since the recall of GT flow clustering is lower than 0.1, all results are clipped away.
For full \nusc\ scores see supplementary material.
}
\vspace{-10mm}
\label{tab:argo_nusc}
\end{table}

\begin{table*}[t]
\centering

\resizebox{\textwidth}{!}{%
 \begin{tabular}{@{}ll|cc|cccc|ccc@{}}
\toprule
  \multicolumn{1}{c}{} &
  \multicolumn{1}{c|}{} &
  \multicolumn{2}{c|}{Movable} &
  \multicolumn{2}{c}{Moving} &
  \multicolumn{2}{c|}{Still} &
  \multicolumn{1}{c}{Vehicle} &
  \multicolumn{1}{c}{Pedestrian} &
  \multicolumn{1}{c}{Cyclist} \\ \cmidrule(l){3-11} 
 &
  \multicolumn{1}{c}{} &
  \multicolumn{2}{|c|}{AP@0.4$\uparrow$} &
  \multicolumn{2}{c}{AP@0.4$\uparrow$} &
  \multicolumn{2}{c|}{AP@0.4$\uparrow$} &
  AP@0.4 $\uparrow$ &
  AP@0.4 $\uparrow$ &
  AP@0.4 $\uparrow$ \\
 &
  \multicolumn{1}{c}{} &
  \multicolumn{1}{|c}{BEV} &
  \multicolumn{1}{c|}{3D} &
  \multicolumn{1}{c}{BEV} &
  \multicolumn{1}{c}{3D} &
  \multicolumn{1}{c}{BEV} &
  \multicolumn{1}{c|}{3D} &
  BEV &
  BEV &
  BEV  \\ \midrule 
 \parbox[t]{3mm}{\multirow{2}{*}{\rotatebox[origin=c]{90}{\textcolor{gray}{Sup.}}}} 
 
    & CP\cite{centerpoint}               & 0.765 & 0.684                    & 0.721 & 0.624 & 0.735 & 0.657 & 0.912 & 0.513 & 0.134   \\ 
    & TF\cite{transfusion}               & 0.746 & 0.723                    & 0.714 & 0.668 & 0.733 & 0.710 & 0.918 & 0.457 & 0.216   \\ \midrule
    \parbox[t]{1mm}{\multirow{2}{*}{\rotatebox[origin=c]{90}{\textcolor{gray}{Unsupervised}}}} 
    & \dbscan \cite{dbscan}          & 0.027 & 0.008                    & 0.009 & 0.000 & 0.027 & 0.006 & 0.184 & 0.002 & 0.001 \\ 
    & \dbscan(SF)      & 0.026 & 0.010                    & 0.064 & 0.041 & 0.000 & 0.000 & 0.073 & 0.010 & 0.009 \\
    & \dbscan(GF)      & 0.114 & 0.071                    & 0.318 & 0.120 & 0.000 & 0.000 & 0.113 & 0.111 & 0.240  \\ %
    & RSF\cite{rsf}              & 0.030 & 0.020                    & 0.080 & 0.055 & 0.000 & 0.000 & 0.109 & 0.000 & 0.002  \\ 
    & SeMoLi \cite{semoli} $\dagger$ & - & 0.195 & - & \textbf{0.575} & - & - & - & - & - \\
    & \textbf{LISO}-CP & 0.292 & 0.211  & 0.272 & 0.204  & 0.208 & 0.140 & 0.607 & 0.029 & 0.010  \\ \midrule
    \parbox[t]{2mm}{\multirow{2}{*}{\rotatebox[origin=c]{90}{\textcolor{gray}{Self Train}}}} 
    & Oyster-CP\cite{oyster}        & 0.217 & 0.084                    & 0.151 & 0.062 & 0.176 & 0.056 & 0.562 & 0.000 & 0.000 \\ %
    & Oyster-TF\cite{oyster}        & 0.121 & 0.015                    & 0.051 & 0.007 & 0.098 & 0.010 & 0.475 & 0.000 & 0.000 \\ %
    & \textbf{LISO}-CP & \textbf{0.380} &  \textbf{0.308} & \textbf{0.350} & 0.296 & \textbf{0.322} & \textbf{0.255} & \textbf{0.695} & \textbf{0.055} & \textbf{0.022} \\ 
    & \textbf{LISO}-TF & 0.327 & 0.208                    & 0.349 & 0.245 & 0.233 & 0.126 & 0.669 & 0.024 & 0.012 \\ 

\bottomrule
\end{tabular}%
}
\caption{\textbf{Evaluation on \wod~dataset}:
We evaluate using the protocol of \cite{semoli,waymo}, using an area of whole 100\si{\m}$\times$40\si{\m} \bev~grid around the ego vehicle, considering objects that move faster than 1\si{\m/\s} to be \textit{moving} (difficulty level L2).
CP, TF: network architecture, in the first two lines trained supervised for comparison.
$\dagger$: Results taken from \cite{semoli}.
SF: \lsf\ by SLIM, GF: ground truth \lsf.
For class-specific 3D detection scores see supplementary material.
}
\vspace{-10mm}
\label{tab:waymo}
\end{table*}

\begin{table}[t]
\centering
\resizebox{\textwidth}{!}{%
\begin{tabular}{@{}l|cccc|cccccc@{}}
\toprule
\multicolumn{1}{c|}{} &
  \multicolumn{4}{c|}{Movable (Moving \& Still)} &
  \multicolumn{2}{c}{Car} &
  \multicolumn{2}{c}{Pedestrian} &
  \multicolumn{2}{c}{Cyclist} \\ \cmidrule{2-11}

\multicolumn{1}{c|}{} &
  \multicolumn{2}{c}{BEV-IOU} &
  \multicolumn{2}{c|}{3D-IOU} &
  \multicolumn{2}{c}{BEV-IOU} &
  \multicolumn{2}{c}{BEV-IOU} &
  \multicolumn{2}{c}{BEV-IOU} \\
  &
  AP@0.3$\uparrow$ &
  AP@0.5$\uparrow$ &
  AP@0.3$\uparrow$ &
  AP@0.5$\uparrow$ &
  AP@0.3$\uparrow$ &
  AP@0.5$\uparrow$ &
  AP@0.3$\uparrow$ &
  AP@0.5$\uparrow$ &
  AP@0.3$\uparrow$ &
  AP@0.5$\uparrow$ \\ \midrule
  CP \cite{centerpoint}              & 0.755 & 0.690 & 0.736 & 0.601 & 0.814 & 0.794 & 0.370 & 0.128 & 0.409 & 0.157 \\ 
  TF \cite{transfusion}              & 0.747 & 0.665 & 0.729 & 0.582 & 0.820 & 0.776 & 0.311 & 0.096 & 0.263  & 0.032 \\ 
\midrule
  \dbscan \cite{dbscan}          & 0.023   & 0.002 & 0.010    & 0.000 & 0.026 & 0.005 & 0.000 & 0.000 & 0.064 & 0.007 \\ 
  RSF \cite{rsf}              & 0.029   & 0.019 & 0.029    & 0.011 & 0.066 &0.049  & 0.000 & 0.000 & 0.192 & 0.043 \\ 
\midrule
Oyster-CP\cite{oyster}        & 0.235   & 0.098 & 0.114    & 0.000 & 0.327 & 0.135 & 0.000 & 0.000 & 0.000 & 0.000 \\  
Oyster-TF\cite{oyster}        & 0.273   & 0.088 & 0.128    & 0.000 & 0.364 & 0.121 & 0.000 & 0.000 & 0.019 & 0.000 \\ 
\textbf{LISO}-CP & \textbf{0.446}   & \textbf{0.330} & \textbf{0.419}  & \textbf{0.159} & \textbf{0.520} & \textbf{0.411} & \textbf{0.097} & \textbf{0.019} & \textbf{0.445} & \textbf{0.053} \\ 
\textbf{LISO}-TF & 0.361   & 0.207 & 0.294    & 0.036 & 0.425 & 0.297 & 0.084 & 0.014 & 0.348 & 0.003 \\ 

\bottomrule
\end{tabular}
}

\caption{\textbf{Evaluation on \kitti~dataset}:
We evaluate on the forward facing field of view where GT annotations are available, but run inference on the whole $100\times100\si{\m}$ \bev~grid.
Also note that flow annotations are not available for \kitti\ Object.
CP, TF: network architecture, in the first two lines trained supervised for comparison.
}
\vspace{-10mm}
\label{tab:kitti}
\end{table}

\subsection{Datasets and Metrics}
We evaluate our method on four different \ad\ datasets.
For a fair comparison, we compute metrics for \textit{movable} objects by mapping all animate objects (Cars, Trucks, Trailers, Motorcycles, Cyclists, Pedestrians and other Vehicles) to a single category and discarding all inanimate objects (Barrier, Traffic Cone,...) since our method does not predict any class attributes.
In \tabref{tab:waymo} and \tabref{tab:kitti} we also give class-based results for completeness. Class labels for true positives are taken from ground truth.
For false positives, the predicted class label is assigned randomly according to the label distribution in the dataset.

\boldparagraph{Waymo Open Dataset (\wod)} \cite{sun2019a} is a large, geographically diverse dataset recorded with a proprietary
high quality lidar.
We evaluate using the protocol of \cite{waymo,semoli}, using an area of whole 100m$\times$40m \bev\ grid around the ego vehicle, artificially cropping the predictions of our method down to this reduced area.

\boldparagraph{\kitti} \cite{geiger2012a, Geiger2013} is recorded using a Velodyne HDL64 \lidar~sensor, where some parts have been annotated with 3D object boxes and tracking information in the forward facing camera field of view.
A large portion of the published data is only available in raw format without annotations, which our method is able to leverage, due to not requiring any ground truth annotations for training.

\boldparagraph{Argoverse 2} \avv\ \cite{Argoverse2} is recorded using two stacked Velodyne VLP32 \lidar~sensors and annotated with 3D object boxes.
On \avv\ and \kitti, we evaluate both 2D (in \bev\ space) and 3D box IoU at IoU thresholds of 0.3 and 0.5 in the area of $100\times100\si{\m}$ around the ego vehicle.

\boldparagraph{\nusc} \cite{caesar2020nuscenes} is recorded using a Velodyne VLP32 \lidar~sensor.
We evaluate using the \nusc~protocol.
Please note that models trained with our method get a high penalty on the Nuscenes Detection Score (NDS), because they cannot distinguish object classes and therefore score an Average Attribute Error of 1.0.

\subsection{Networks}
We evaluate our proposed method with two \sota\ \lidar\ object detection networks of different architecture: Centerpoint \cite{centerpoint} and Transfusion-L \cite{transfusion}.
For both we do not make any modification to the published implementation or network architecture or losses.
We train both networks with a batch size of four.

\boldparagraph{CenterPoint}
\cite{centerpoint} is based on 2D convolutions in \bev\ space.
Object centers are represented as heatmap in the \bev, which are then reduced to a sensible amount of boxes using non-maximum-suppression.

\boldparagraph{Transfusion-L}
\cite{transfusion} is a Transformer based architecture, which is applied after initial encoding of the LiDAR point cloud into a 2D \bev~feature tensor. 

\subsection{Baselines}
\label{subsec:baselines}
Besides the obvious baseline of training the object detection networks supervised from scratch on ground truth from the dataset, we also compare to multiple strong unsupervised baselines. For details, see Sec.~\ref{sec:related}.

\boldparagraph{RSF}
We include RSF~\cite{rsf} in our evaluation as strong representative of methods doing object distilling from motion cues.
We ran experiments using the published code of the authors.

\boldparagraph{\dbscan}
This algorithm \cite{dbscan} clusters points in the point cloud with similar 3D locations into objects. We additionally evaluate extending the cluster space to 6D by either using SLIM scene flow (SF) or ground-truth scene flow (GT). Thus, the "DBSCAN(SF)" baseline essentially corresponds to the raw detections used for our initial \pgt\ generation. We use the implementation from \cite{scikit-learn} with parameter values 1.0 for epsilon and 5 for the minimum number of points for a valid cluster.

\boldparagraph{SeMoLi}
\cite{semoli} is the most recent baseline.
As it showed to outperform \cite{waymo} as self-supervised object detection method we include only published results of \cite{semoli} in our evaluation.

\boldparagraph{Oyster}
We reimplemented \cite{oyster} using PyTorch \cite{pytorch}, and apply the proposed framework to the used networks CenterPoint and Transfusion-L (abbreviated Oyster-CP and Oyster-TF) to be as close as possible to our method and evaluation.
We verify our re-implementation, by reproducing the reported metrics by \cite{oyster}\ on \avv, see \tabref{tab:argo_nusc}.

\subsection{Results}
\boldparagraph{Quantitative Results}
On all four datasets we see that our method consistently outperforms all self-supervised baselines in all metrics for \textit{movable} objects, see \tabref{tab:argo_nusc}, \tabref{tab:waymo}, and \tabref{tab:kitti}.
Only SeMoLi\cite{semoli} beats LISO on \textit{moving} objects (see \tabref{tab:waymo}) but significantly drops below LISO's performance on \textit{movable} objects, hinting to difficulties with generalization.
Our method does not suffer from this performance gap and performs nearly equally well on \textit{movable} and \textit{moving} objects.
We also observe that, despite mostly better performance in the supervised case, Transfusion-L responds less favorable than Centerpoint to the \ssup\ training approaches.
We suspect that Centerpoint, due to its convolutional architecture and its lack of positional encoding compared to Transfusion-L, is less susceptible to overfitting to where \textit{moving} objects have been observed and where not to expect them.

We also observe that the \nusc\ dataset seems to be the most challenging for the \ssup\ methods, leading to the biggest gap between supervised and unsupervised object detection performance.
We attribute this to the sparseness of the lidar sensor used in the dataset with only 32 layers.
Common to all \ssup\ methods is the difficulty in estimating the forward orientation of objects (AOE on \nusc, $2\pi$ periodic, see \tabref{tab:argo_nusc}), which the pure IoU based metrics cannot reveal.

\begin{figure}
    \centering
    \includegraphics[width=0.46\textwidth, clip]{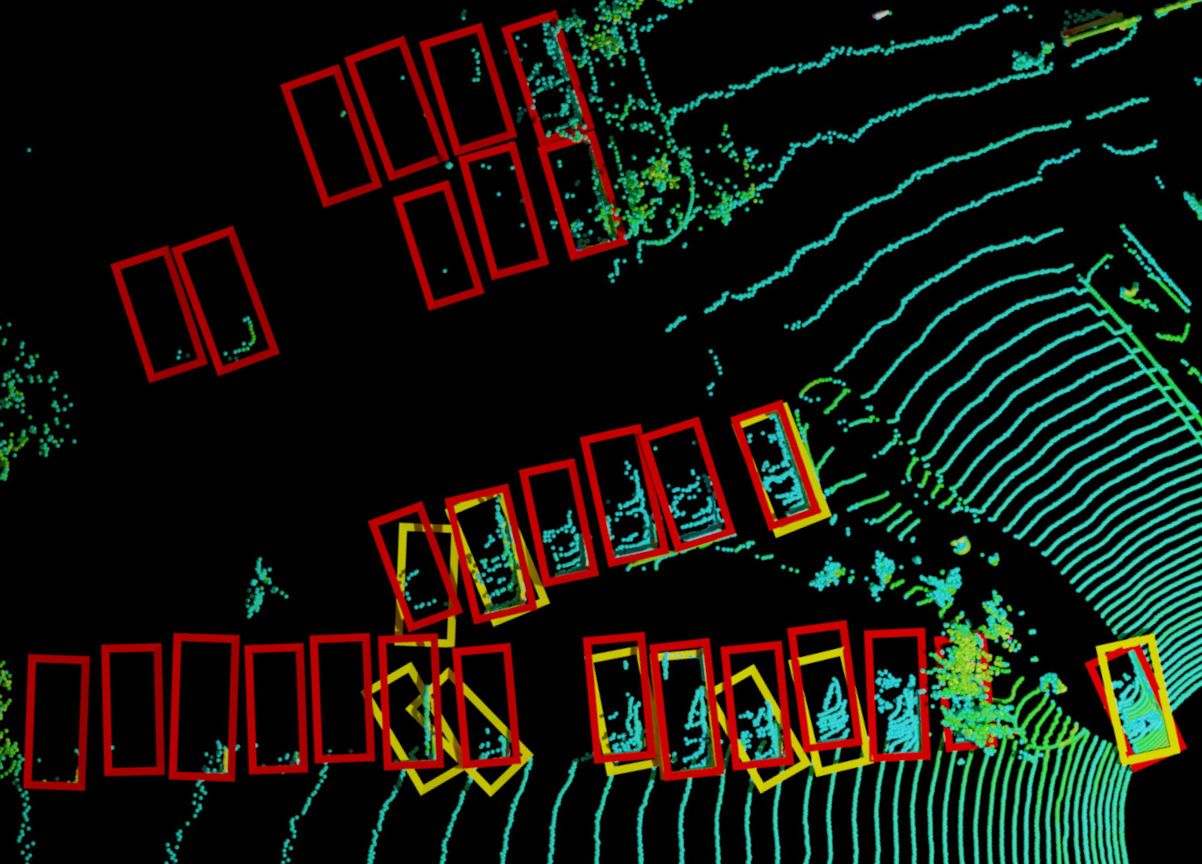}
    \caption{
    Missed ground truth boxes in the \wod\ dataset. 
    }
    \vspace{-3mm}
    \label{fig:waymogt}
\end{figure}

\boldparagraph{Qualitative Results}
In \figref{fig:qualitative} we compare our method to the best-performing baseline, Oyster, using Centerpoint as network architecture.
It can be seen, that our method has higher accuracy and especially predicts box orientations more accurately.

Fig.~\ref{fig:waymogt}, in contrast, showcases an important failure-case and general problem of all unsupervised methods: ground truth, especially in the WOD dataset, is also annotated in very sparse and distant regions. It is very challenging to generate \pgt\ for such regions whereas supervised trainings do not have to generalize from \textit{moving} to such \textit{movable} (but very sparse) objects. This may be the main reason for a still noticeable gap in evaluation results between supervised and unsupervised methods.

\subsection{Ablation Study}
We investigated the influence of different choices for the components in our pipeline, see \tabref{tab:ablation} and cf. Fig.~\ref{fig:overview}.

\boldparagraph{Motion Cues as Clustering Input} When adding motion cues to box clustering for Oyster we observe that detection performance already increases noticeably (comparing \#3 to \#4, \tabref{tab:ablation}).
This demonstrates that self-supervised motion cues are a key ingredient to get high-quality initial \pgt\ as opposed to just using clustered 3D points.

\begin{table}[b]
\vspace{-5mm}
\centering
\begin{tabular}{@{}lllcccccc@{}}
\toprule
  Tag &
  Cluster Input & 
  Method &
  Self Train &
  Track Optim. &
  AP@0.4(BEV) &
  AP@0.4(3D) \\
  \midrule
\#1 & P, SF   &  no tracker          &             &                    & 0.177  & 0.086 \\ %
\#2 & P, SF   &  no tracker          & \checkmark  &                    & 0.201  & 0.131 \\ %
\#3 & P       & Oyster               & \checkmark  &                    & 0.217  & 0.084 \\ 
\#4 & P, SF   & Oyster               & \checkmark  &                    & 0.255  & 0.104 \\ %
\#5 & P, SF   & LISO(K, SF)         &              &                    & 0.279  & 0.209 \\ %
\#6 & P, SF   & LISO(K, SF)         & \checkmark   &                    & 0.360  & 0.290 \\ %
\#7 & P, SF   & LISO(K, SF)         &              & \checkmark         & 0.292  & 0.211  \\  %
\#8 & P, SF   & LISO(K, SF)         & \checkmark   & \checkmark         & \textbf{0.380} & \textbf{0.308} \\ \midrule
\#9 & P, SF   & LISO(G, SF)         &\checkmark    & \checkmark         & 0.411 & 0.339\\
\#10 & P, GF   & LISO(G, GF)         & \checkmark   & \checkmark         & 0.423 & 0.349  \\ %

\bottomrule
\end{tabular}%
\caption{\textbf{Ablation study of different parts of our pipeline} on \wod\ with Centerpoint~\cite{centerpoint}.
\textbf{Abbreviations:} P: points, SF: \ssup\ \lsf\ (SLIM), GF: ground truth \lsf. K: KISS-ICP ego-motion, G: ground truth ego-motion. 
}
\vspace{-5mm}
\label{tab:ablation}
\end{table}

\boldparagraph{Motion Cues for Tracking} Our \lsf-based tracker is the biggest contributing factor to the success of our method: the ablation reveals that leveraging motion cues greatly improves tracking (comparing \#4 to \#6) and thus ultimately improves self-training, i.e. more accurate tracking allows for much more strictness when matching and filtering tracklets. (The Oyster tracker uses a matching threshold of $5\si{\m}$, our threshold is only $1.5\si{\m}$). This strictness leads to higher quality \pgt.

\boldparagraph{Effect of Self Training} Even without using any tracker to filter tracklets self training has a benefit (comparing \#1 to \#2).
This is due to the network generalizing to new instances and confidence thresholding the detections.
However, having an additional way to discard implausible detections (by tracking) amplifies the positive effect of self training (comparing \#5 to \#6 and \#7 to \#8), as it can better prevent false positives from entering the \pgt.

\boldparagraph{Track Optimization} The added benefit of Track Optimization is more independent of Self Training (comparing going from \#5 to \#7 with going from \#6 to \#8). 
Finally we notice, that with the combination of a strict tracker, track optimization and self training, the performance is very close to using ground truth flow and ego-motion (comparing \#8 to \#9 and \#10).

\boldparagraph{Hyperparameters for Iterative Training}
\figref{fig:pgtNetQuality} demonstrates that the performance of our proposed method is relatively consistent across a range of hyperparameters.
However, using too few steps between regeneration of \pgt\ ($s=20k$) leads to degrading performance, because the network does not have enough time to generalize and stabilize using the \pgt, which then has negative effects during regeneration of the \pgt.
The experiment reveals that doing multiple iterations of self-learning and dropping network-weights to allow the network to adjust is beneficial for a good performance. We selected $r=2$ and $s=30k$ as a good compromise between performance and speed for all other experiments.

\section{Conclusion}
\label{sec:conclusion}
We have proposed a novel framework for \ssup\ 3D lidar object detection based on \ssup\ \lsf.
Using simple clustering on \lsf\ in combination with a novel flow-based tracker, we were able to generate \pgt\ with \textit{moving} objects at high precision.
This enabled us to bootstrap \sota\ 3D lidar object detectors, without using any human labels.
The detector generalizes from detecting \textit{moving} to detecting \textit{movable} objects over multiple self-training iterations.
Experiments revealed that our trajectory-regularized self-learning, based on our scene flow-based tracker, is key to the success of our method.
We have demonstrated the effectiveness of our approach on two \sota\ lidar object detectors as well as four \ad\ datasets. All experiments were conducted with the same set of hyperparameters, demonstrating the robustness of our method.
Our method achieves a significant improvement in the state-of-the-art of \ssup\ lidar object detection on all four datasets.
Code will be released.

The biggest limitation of our approach is that our method is not able to distinguish different classes.
Future work could be around generating \pgt\ for class labels, e.g. based on motion or size characteristics.

\FloatBarrier

\bibliographystyle{splncs04}
\bibliography{main}
\author{}
\institute{}
\clearpage
\appendix 
\setcounter{page}{1}

\title{Supplementary Material}
\maketitle

\section{Implementation Details}
We run the networks Centerpoint~\cite{centerpoint} and Transfusion-L~\cite{transfusion} on $100\times100\si{\m}$ \bev~grids around the ego vehicle.
We use non-maximum suppression with a threshold of $0.1$ (2D \bev\ IoU) for the detections.
The optimizers, as well as their learning rate schedules are kept from the respective original implementations, but the schedules are shortened to match the lifecycle of the network weights during the iterative rounds of self-training.
For the zero-shot generalization required by Oyster~\cite{oyster} after the first round, we found that starting from an initial \bev\ range of $50\times50\si{\m}$, and then extending to $100\times100\si{\m}$, gave the best results.
For \dbscan\ we used $\varepsilon=1.0$, and $\text{minPts}=5$.
We optimize all tracks using Adam optimizer with learning rate 0.1 for 2000 steps for a complete point cloud sequence batched (at the same time), which takes less than 2\si{\s} per \nusc\ session on a Nvidia V100 GPU.

\section{Self-supervised \lsf}
\begin{table}[b]
    \centering
    \begin{tabular}{@{}llll@{}}
    \toprule
    Train Data & Val Data &AEE(moving)[m] \(\downarrow\) & AEE(static)[m]\(\downarrow\) \\
    \midrule
    \avv~Train& \avv~Val & 0.075 & 0.079 \\
    \kitti~Raw & \kitti~Tracking  & 0.092 & 0.104 \\
    \nusc~Train & \nusc~Val  & 0.132 & 0.077 \\
    \wod~Train & \wod~Val  & 0.091 & 0.085 \\
    \bottomrule
    \end{tabular}%
    \caption{\textbf{Lidar scene flow metrics of SLIM \cite{slim} on the datasets (evaluated on val split), for a \bev~range of $\mathbf{120\times120\si{\mathbf{\m}}}$.} Note that for \kitti, we only evaluate the forward-facing \fov~which has been annotated with tracked objects. Objects faster than $1\si{\m /\s}$ are considered \textit{moving}.
    AEE refers to the average endpoint error across either all moving or static points.
    }
\label{tab:slim}
\end{table}

As mentioned in Sec.~\ref{sec:eval}, we extend the \bev\ range of SLIM~\cite{slim} from $70\times70\si{\m}$, $640\times640$ pixels to $120\times120\si{\m}$ and $920\times920$ pixels, but make no further modifications to the network.
This results in the \sota\ scene flow quality described in \tabref{tab:slim}.

The small performance gap of our method between using ground truth and SLIM \lsf\ (comparing the last and the second-to-last row of \tabref{tab:ablation}) demonstrates that SLIM \lsf\ has suitable quality for our method, and also that our method does not require absolutely perfect \lsf\ estimates to work well.
Ground truth \lsf\ is generated using the recorded vehicle egomotion for static points and the tracking information (bounding boxes) of moving objects.

\section{Performance of \lsf\ clustering on \nusc}
In the evaluation on \nusc\ (see \tabref{tab:nusc_full}), the worse performance of using \dbscan~\cite{dbscan} clustering on ground truth \lsf\ compared to using \dbscan\ on SLIM \lsf\ is surprising.
\begin{figure}
\resizebox{0.99\linewidth}{!}{%
\includegraphics[height=4cm]{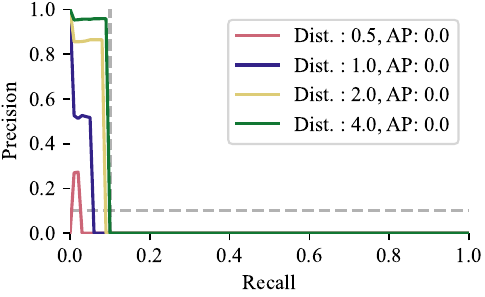}
\hspace{-0.5mm}
\includegraphics[height=4cm]{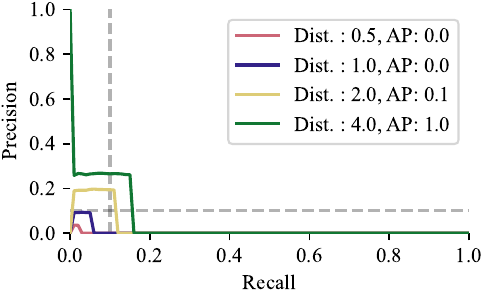} 
}
\caption{
\textbf{Performance comparison of clustering ground truth \lsf (top) and SLIM~\cite{slim}\ \lsf\ (bottom) on the \nusc\ dataset.}
The methods are evaluated according to the official \nusc\ protocol on the validation split.
The dashed line represents the minimum threshold for precision and recall of $0.1$, all results below these two thresholds are discarded.
This leads to the surprising effect that the AP score is higher when using SLIM \lsf, but this is only a result of the clipping dictated by the \nusc\ evaluation protocol. 
}
\label{fig:flow_clustering}
\end{figure}
However, this peculiar effect is explained by \figref{fig:flow_clustering}, which shows the full precision-recall curves, generated using the official \nusc\ protocol on the validation split \cite{caesar2020nuscenes}.
The \nusc\ protocol uses minimum precision and recall value thresholds of $0.1$, discarding all results below these thresholds.
As mentioned in \secref{sec:pgtgeneration}, we assign confidence score of $1.0$ to all clusters discovered by \dbscan.
This causes all detections generated using \dbscan\ on ground truth \lsf\ to be discarded.

\section{Quality of \pgt\ during Iterative Self-Training}
\begin{figure*}
\includegraphics[width=\textwidth]{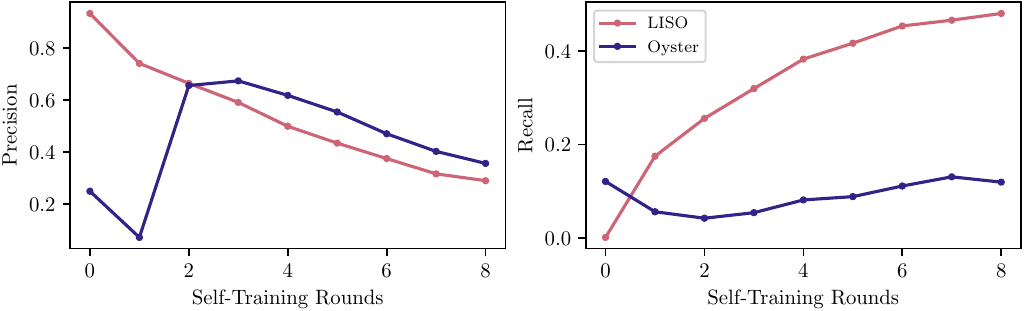} 
\caption{
\textbf{Precision and recall of the (tracked) \pgt\ generated by Oyster and LISO over the course of self-training of Centerpoint on \wod\ (training split).}
Precision and recall are computed like in the AP metrics used in \figref{fig:pgtNetQuality} and \tabref{tab:waymo}, i.e. true positives are occurences where the \bev\ IoU between ground truth and predicted boxes is greater than 0.4, but at a specific confidence threshold:
For Oyster, we use the reported value from the publication $c=0.4$ \cite{oyster}.
For LISO, we use $c=0.3$ and only discard the learned weights every other round, as stated in \secref{sec:pgtgeneration}.
Note that the dip in Oyster's performance at round 1 stems from the zero-shot generalization, where the network is tasked to generalize from the training on the initial \pgt\ generated on the smaller \bev\ range to the full, previously unseen \bev\ range, going from $50\times50\si{\m}$ to $100\times100\si{\m}$.
}
\label{fig:pgt_time}
\end{figure*}

\begin{figure}
    \centering
    \includegraphics[width=.49\columnwidth, trim=0 0 0 0cm, clip]{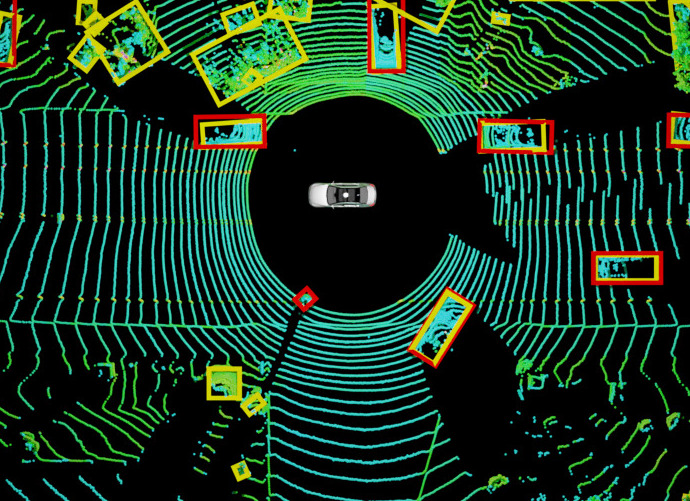}
    \includegraphics[width=.49\columnwidth, trim=0cm 0cm 0 0cm, clip]{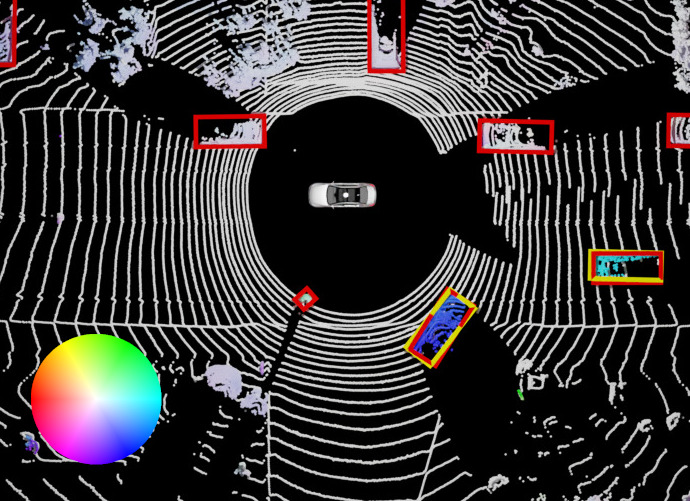}
    \caption{
    \textbf{Clustering results for the initial \pgt\ generation on \wod.} \red{Red} boxes are ground truth boxes, \yellow{yellow} are predictions. 
    \textbf{Left: Oyster} Clustering result on points, with high recall but low precision.
    \textbf{Right: LISO} Clustering result on points and SLIM \lsf, resulting in in high precision \pgt\ (LISO). Points are colored according to flow direction and magnitude.
    }
    \label{fig:clustering}
\end{figure}

One critical aspect of iterative self-training is the quality of \pgt\ on the training dynamics, as depicted in \figref{fig:pgt_time}.
Finding the right balance between precision and recall in the \pgt\ is crucial for achieving optimal performance during self-training iterations:
In our experiments, we find that having initially a small subset of high precision training samples is superior to having a larger set with higher recall but worse precision, because it allows the model to learn from a smaller but more reliable set of labeled data. 
A larger set of \pgt\ that is collected with less rigorous clustering, tracking and filtering, includes more noisy and mislabeled data.
As discussed in \cite{oyster, modest}, the limited model capacity does prevent the model from overfitting to the inconsistent noises in the \pgt\ to some extent and the model generalizes mostly to the objects of interest \cite{oyster,modest}, but in our experiments, higher quality \pgt\ with less noise ultimately leads to better performance.
Motion cues (i.e. egomotion and \lsf) are the superior clustering and tracking input signal, allowing our method to generate much cleaner initial \pgt\ when compared to Oyster, which we also demonstrate in our ablation in \tabref{tab:ablation}.
\figref{fig:clustering} additionally visually demonstrates the difference between using \lsf\ for initial \pgt\ creation and just using point clouds (Oyster) on an example point cloud: As expected, using \lsf\ leads to fewer false positives in the initial \pgt.

\section{Qualitative Results}
\begin{figure*}
    \centering
    \includegraphics[width=0.49\textwidth, trim=0 0 0 11cm, clip]{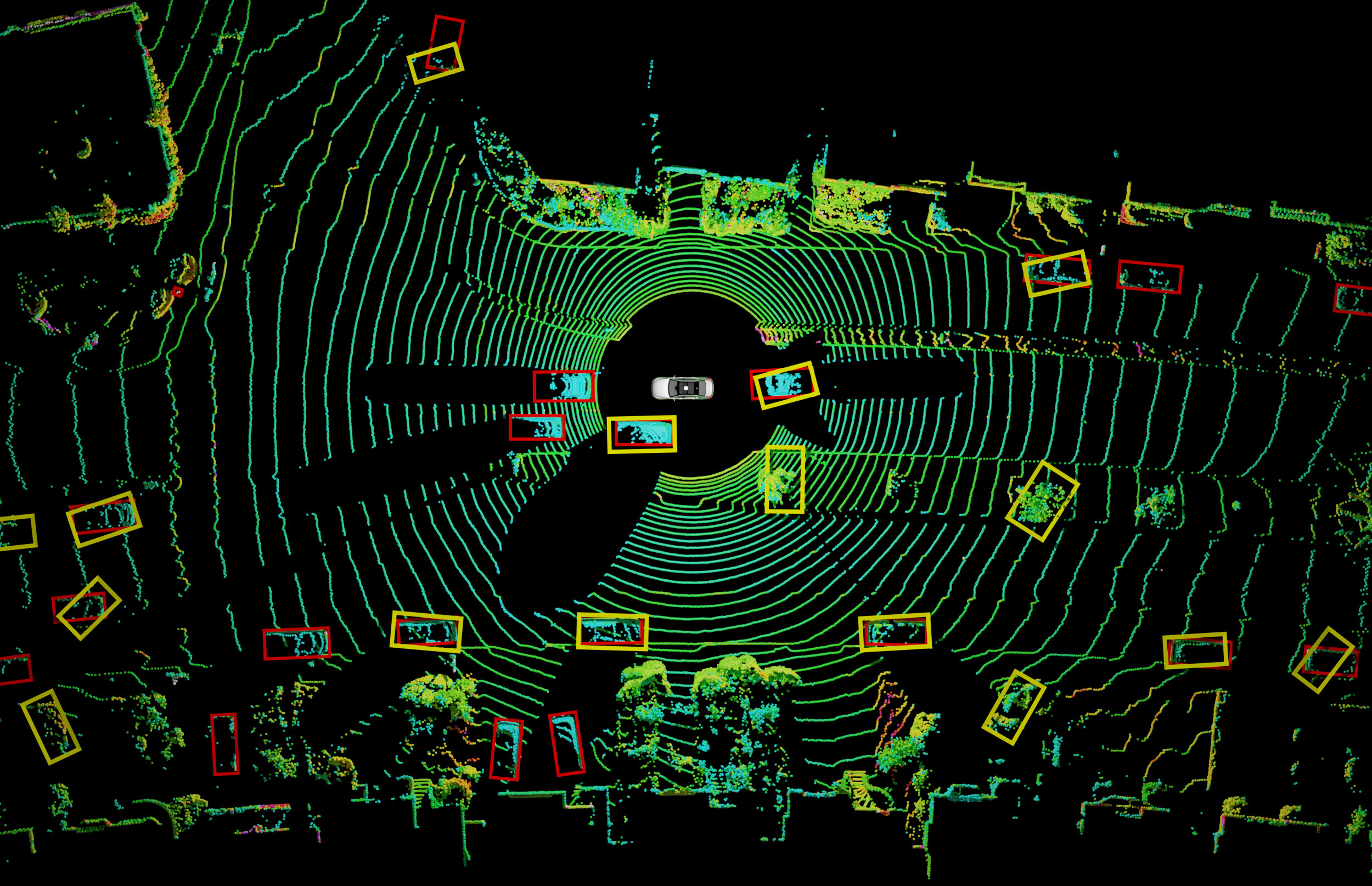}
    \includegraphics[width=0.49\textwidth, trim=0 0 0 11cm, clip]{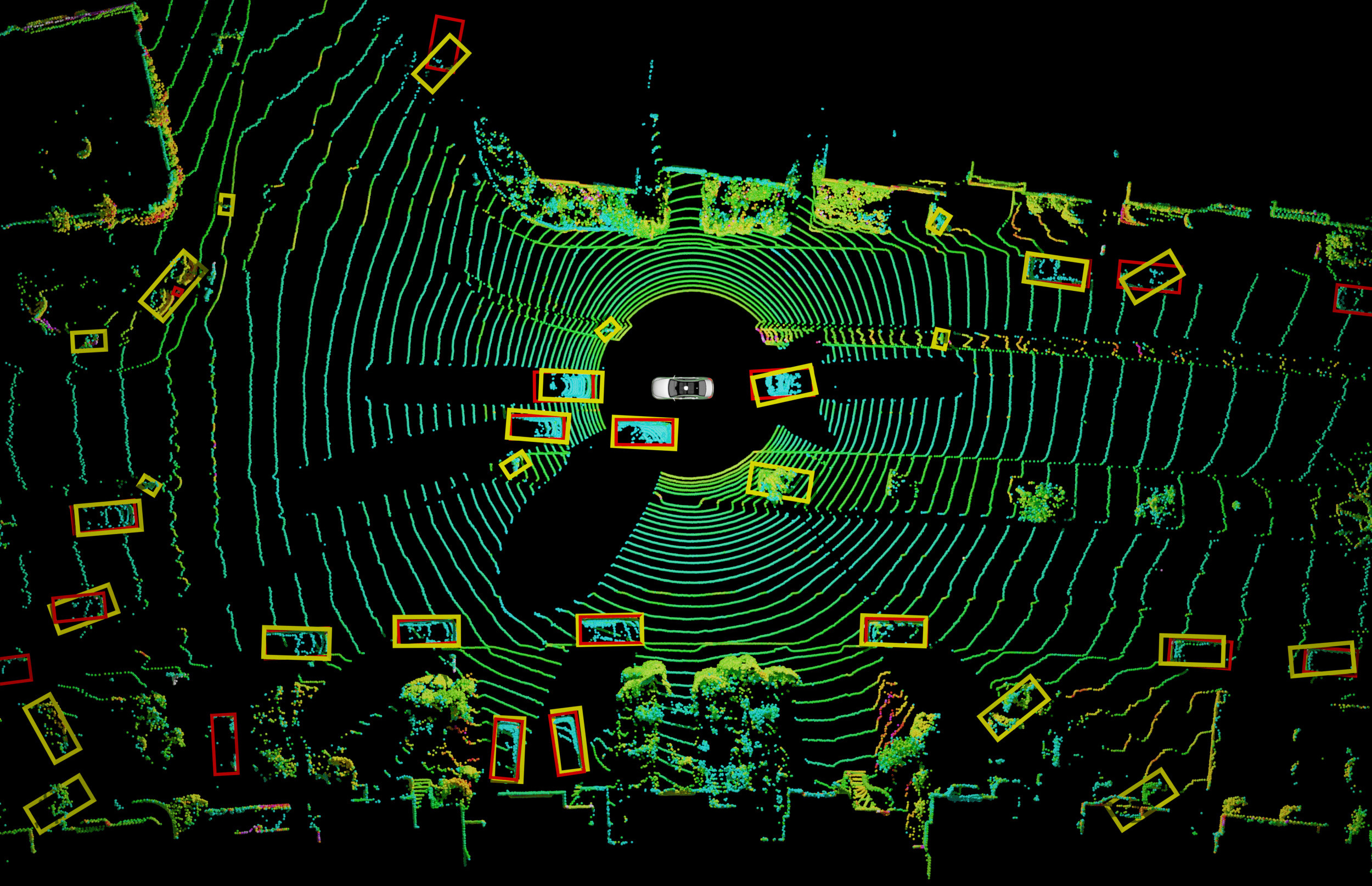}
    \\
    \includegraphics[width=0.49\textwidth, trim=6cm 5cm 0 6cm, clip]{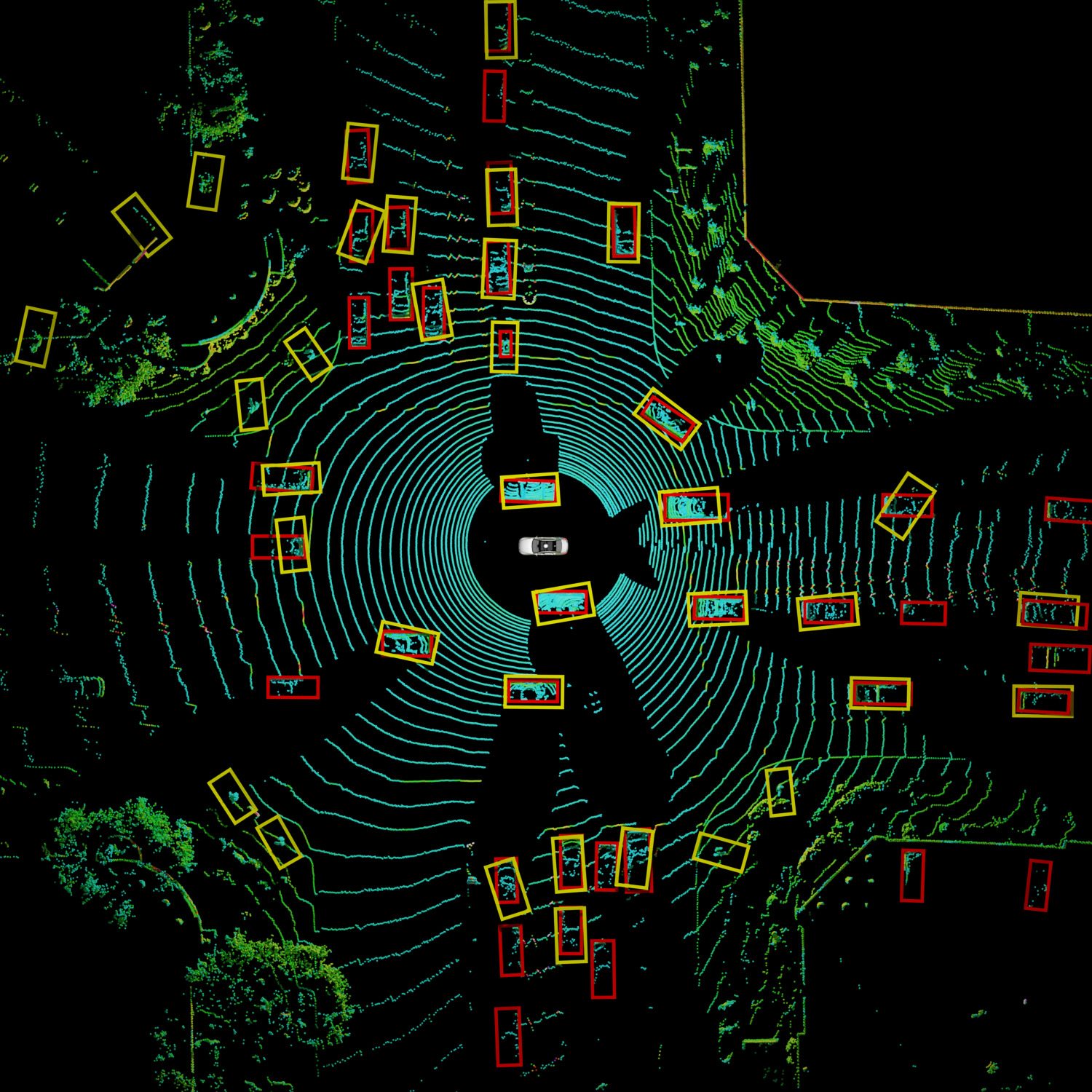}
    \includegraphics[width=0.49\textwidth, trim=6cm 5cm 0 6cm, clip]{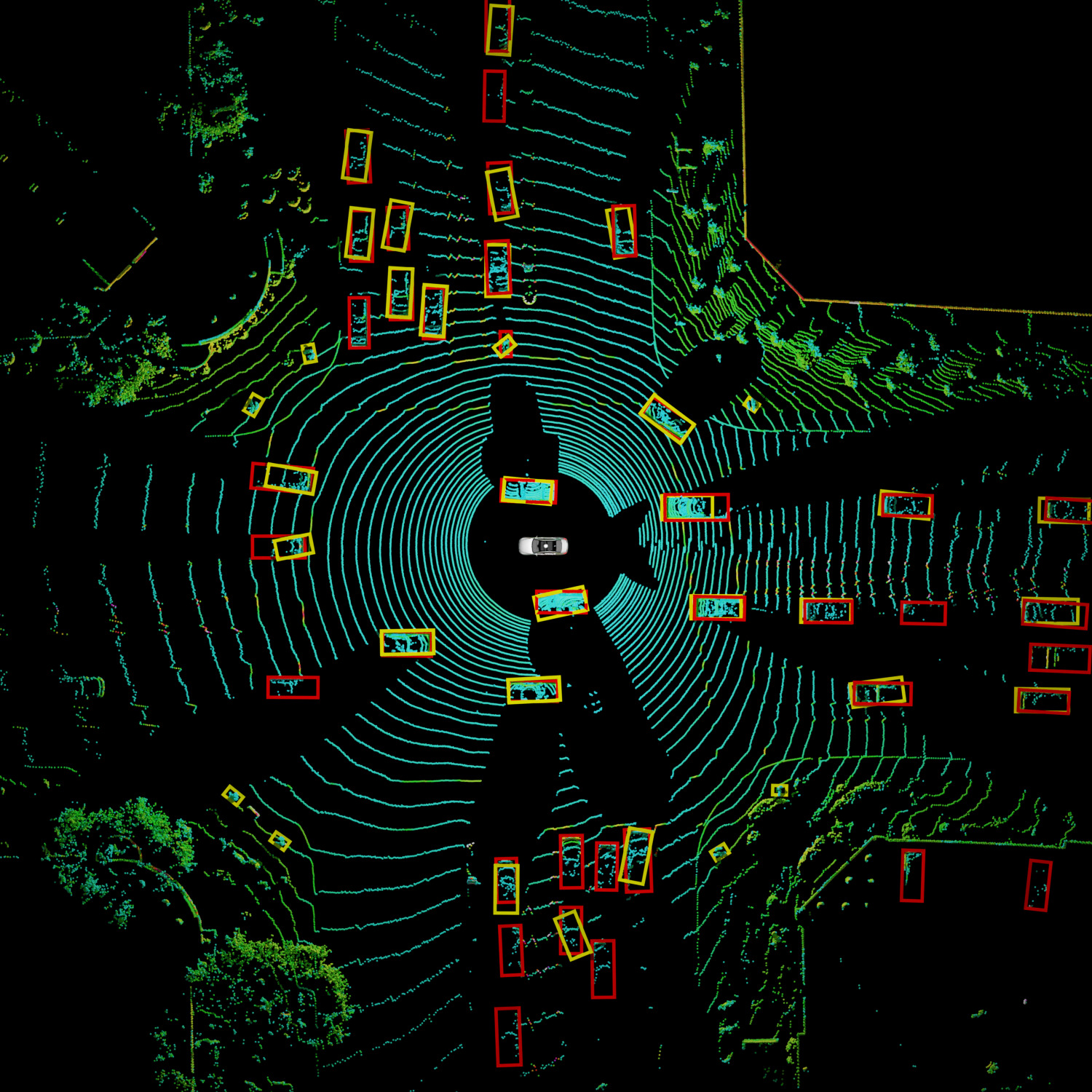}
    \\
    \includegraphics[width=0.49\textwidth, trim=0cm 4cm 10cm 5cm, clip]{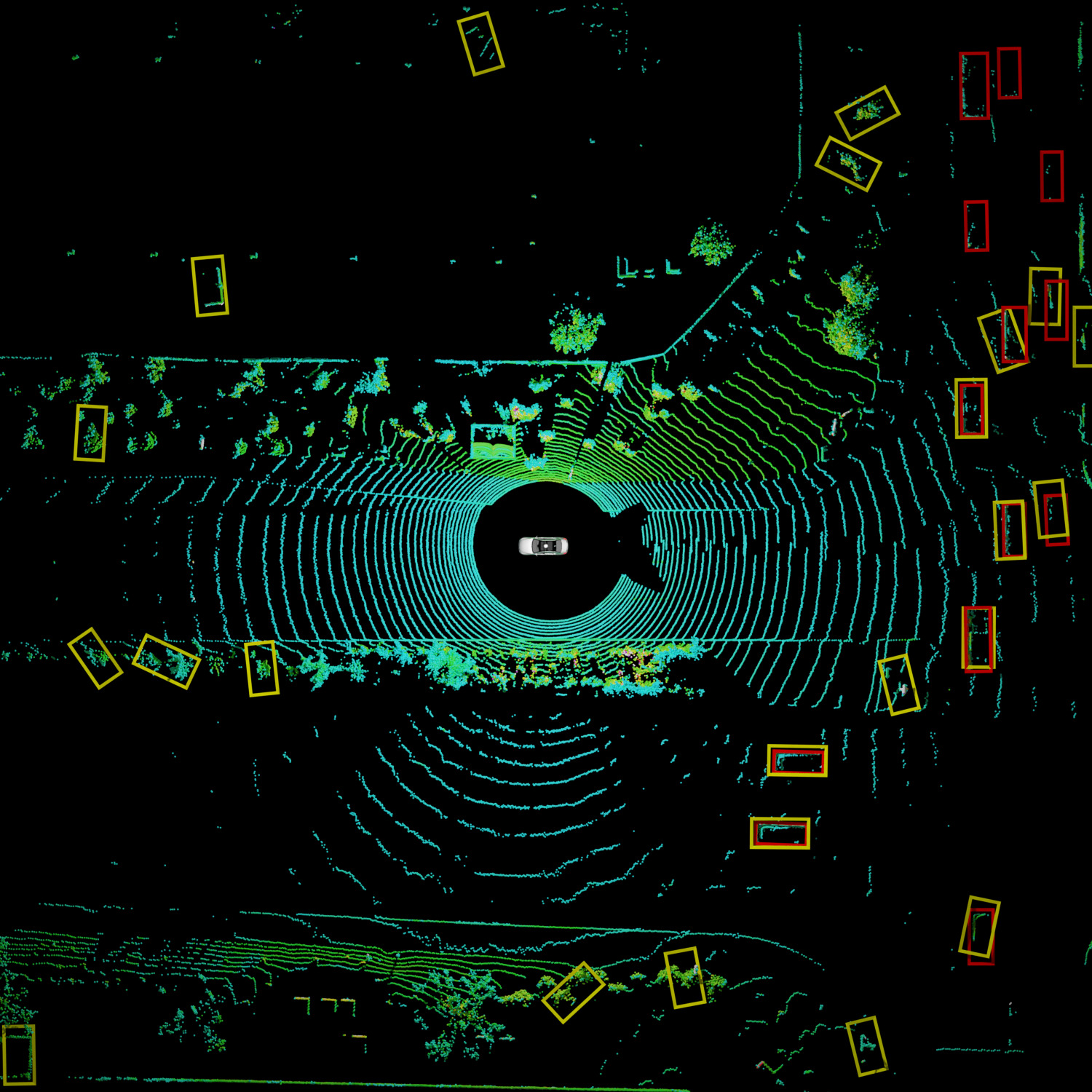}
    \includegraphics[width=0.49\textwidth, trim=0cm 4cm 10cm 5cm, clip]{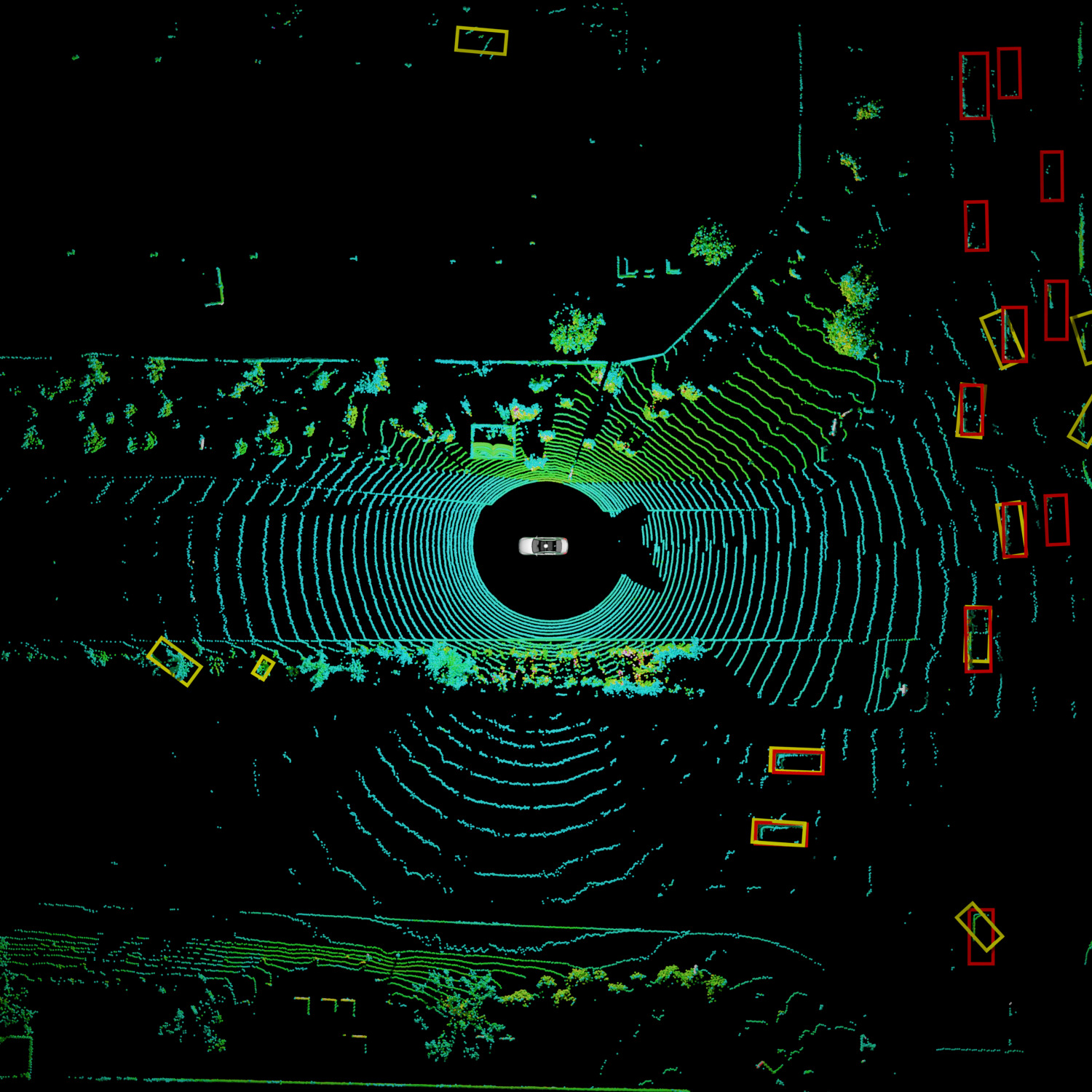}
    \caption{
    \textbf{Qualitative Results on \wod.} \red{Red} boxes are ground truth boxes, \yellow{yellow} are predictions. 
    \textbf{Left:} OYSTER-CP
    \textbf{Right:} LISO-CP
    Both methods struggle to some extent with false positive detections, but Oyster much more so, despite using the higher confidence threshold.
    We attribute this to the fact that Oyster has noisier initial \pgt, which leads to wrong training signals.
    }
    \label{fig:qualitative2}
\end{figure*}

For more qualitative comparisons besides  \figref{fig:qualitative2} or \figref{fig:qualitative}, we kindly refer the reader to the video accompanying this supplement.

\section{Quantitative Results}
In \tabref{tab:waymo_full} and \tabref{tab:nusc_full} we have more detailed metrics for \wod\ and \nusc. 
Please note that models trained with our method get a high penalty on the Nuscenes Detection Score (NDS), because they cannot distinguish object classes and therefore score an Average Attribute Error of 1.0.
\begin{table*}[t]
\centering
\begin{tabular}{@{}llcccccc@{}}
\toprule
& method      & \begin{tabular}[c]{@{}l@{}}AP$\uparrow$ \\ \end{tabular} & \begin{tabular}[c]{@{}l@{}}NDS$\uparrow$ \end{tabular} & \begin{tabular}[c]{@{}l@{}}ATE$\downarrow$ \end{tabular} & \begin{tabular}[c]{@{}l@{}}AOE $\downarrow$ \end{tabular} & \begin{tabular}[c]{@{}l@{}}ASE$\downarrow$\\ \end{tabular} \\ \midrule
\parbox[t]{3mm}{\multirow{2}{*}{\rotatebox[origin=c]{90}{\textcolor{gray}{GT}}}} 
& CP\cite{centerpoint}          & 0.484  & 0.524  & 0.357 & 0.560 & 0.263 \\ %
& TF\cite{transfusion}          & 0.627  & 0.614  & 0.287 & 0.501 & 0.207 \\ \midrule %
\parbox[t]{3mm}{\multirow{4}{*}{\rotatebox[origin=c]{90}{\textcolor{gray}{Unsup.}}}} 
& \dbscan \cite{dbscan}                      & 0.008  & 0.109  & 0.987 & 3.120  & 0.962 \\ %
& \dbscan(SF)                   & 0.003  & 0.106  & 1.186 & 2.623 & 0.952 \\ %
& \dbscan(GF)                   & 0.000  & 0.000  & 1.000 & 1.0    & 1.0    \\ %
& RSF\cite{rsf}                 & 0.019  & 0.183  & 0.774 & 1.003  & 0.507  \\ %
\midrule
\parbox[t]{3mm}{\multirow{4}{*}{\rotatebox[origin=c]{90}{\textcolor{gray}{Self Train}}}} 
& Oyster-CP\cite{oyster}        & 0.091  & 0.215  & 0.784 & 1.514  & 0.521  \\
& Oyster-TF\cite{oyster}        & 0.093  & 0.233  & 0.708 & 1.564  & 0.448  \\
& \textbf{LISO}-CP              & 0.109  & 0.224  & 0.750 & 1.062 & 0.409   \\ %
& \textbf{LISO}-TF              & \textbf{0.134}   & \textbf{0.270} & \textbf{0.628} & \textbf{0.938} & \textbf{0.408} \\ 

\bottomrule
\end{tabular}
\caption{\textbf{Full evaluation results on \nusc~ dataset}:
We compare \textbf{LISO} with two different network architectures (TF~\cite{transfusion}, CP~\cite{centerpoint}) against different baselines and also give supervised training results as reference (two top lines). Along the AP score we report the \nusc\ detection score NDS, which is a combination of the AP score, average translation, orientation, scale, attribute error/score (ATE, AOE, ASE, AEE respectively).
Note that \nusc\ uses a minimum precision and recall threshold of 0.1, and since the recall of GT flow clustering is lower than 0.1, all results are clipped away.
SF: \lsf\ by SLIM, GF: ground truth \lsf.
}
\label{tab:nusc_full}
\end{table*}

\begin{table*}[t]
\centering

\resizebox{\textwidth}{!}{%
 \begin{tabular}{@{}ll|cc|cccc|cccccc@{}}
\toprule
  \multicolumn{1}{c}{} &
  \multicolumn{1}{c|}{} &
  \multicolumn{2}{c|}{Movable} &
  \multicolumn{2}{c}{Moving} &
  \multicolumn{2}{c|}{Still} &
  \multicolumn{2}{c}{Vehicle} &
  \multicolumn{2}{c}{Pedestrian} &
  \multicolumn{2}{c}{Cyclist} \\ \cmidrule(l){3-14} 
 &
  \multicolumn{1}{c}{} &
  \multicolumn{2}{|c|}{AP@0.4} &
  \multicolumn{2}{c}{AP@0.4} &
  \multicolumn{2}{c|}{AP@0.4} &
  \multicolumn{2}{c}{AP@0.4} &
  \multicolumn{2}{c}{AP@0.4} &
  \multicolumn{2}{c}{AP@0.4} \\
 &
  \multicolumn{1}{c}{} &
  \multicolumn{1}{|c}{BEV} &
  \multicolumn{1}{c|}{3D} &
  \multicolumn{1}{c}{BEV} &
  \multicolumn{1}{c}{3D} &
  \multicolumn{1}{c}{BEV} &
  \multicolumn{1}{c|}{3D} &
  \multicolumn{1}{c}{BEV} &
  \multicolumn{1}{c}{3D} &
  \multicolumn{1}{c}{BEV} &
  \multicolumn{1}{c}{3D} &
  \multicolumn{1}{c}{BEV} &
  \multicolumn{1}{c}{3D} \\ \midrule 
 \parbox[t]{3mm}{\multirow{2}{*}{\rotatebox[origin=c]{90}{\textcolor{gray}{GT}}}} 
 
    & CP\cite{centerpoint}               & 0.765 & 0.684                    & 0.721 & 0.624 & 0.735 & 0.657 & 0.912 & 0.841 & 0.513 & 0.413 & 0.134 & 0.094  \\ 
    & TF\cite{transfusion}               & 0.746 & 0.723                    & 0.714 & 0.668 & 0.733 & 0.710 & 0.918 & 0.899 & 0.457 & 0.429 & 0.216 & 0.187  \\ \midrule
    \parbox[t]{2mm}{\multirow{2}{*}{\rotatebox[origin=c]{90}{\textcolor{gray}{Unsupervised}}}} 
    & \dbscan \cite{dbscan}          & 0.027 & 0.008                    & 0.009 & 0.000 & 0.027 & 0.006 & 0.184 & 0.048 & 0.002 & 0.000 & 0.001 & 0.000 \\ 
    & \dbscan(SF)      & 0.026 & 0.010                    & 0.064 & 0.041 & 0.000 & 0.000 & 0.073 & 0.046 & 0.010 & 0.006 & 0.009 & 0.006 \\
    & \dbscan(GF)      & 0.114 & 0.071                    & 0.318 & 0.120 & 0.000 & 0.000 & 0.113 & 0.075 & 0.111 & 0.063 & 0.240 & 0.151 \\ %
    & RSF\cite{rsf}              & 0.030 & 0.020                    & 0.080 & 0.055 & 0.000 & 0.000 & 0.109 & 0.074 & 0.000 & 0.000 & 0.002 & 0.000   \\ 
    & SeMoLi \cite{semoli} $\dagger$ & - & 0.195 & - & \textbf{0.575} & - & - & - & - & - & - & - & - \\
    & \textbf{LISO}-CP & 0.292 & 0.211  & 0.272 & 0.204  & 0.208 & 0.140 & 0.607 & 0.440 & 0.029 & 0.009 & 0.010 & 0.004  \\ \midrule
    \parbox[t]{2mm}{\multirow{2}{*}{\rotatebox[origin=c]{90}{\textcolor{gray}{Self Train}}}} 
    & Oyster-CP\cite{oyster}        & 0.217 & 0.084                    & 0.151 & 0.062 & 0.176 & 0.056 & 0.562 & 0.204 & 0.000 & 0.000 & 0.000 & 0.000 \\ %
    & Oyster-TF\cite{oyster}        & 0.121 & 0.015                    & 0.051 & 0.007 & 0.098 & 0.010 & 0.475 & 0.058 & 0.000 & 0.000 & 0.000 & 0.000 \\ %
    & \textbf{LISO}-CP & \textbf{0.380} &  \textbf{0.308} & \textbf{0.350} & 0.296 & \textbf{0.322} & |\textbf{0.255} & \textbf{0.695} & \textbf{0.543} & \textbf{0.055} & \textbf{0.037} & \textbf{0.022} & \textbf{0.016}  \\ 
    & \textbf{LISO}-TF & 0.327 & 0.208                    & 0.349 & 0.245 & 0.233 & 0.126 & 0.669 & 0.408 & 0.024 & 0.008 & 0.012 & 0.005  \\ 

\bottomrule
\end{tabular}%
}
\caption{\textbf{Full evaluation results on \wod~dataset}:
We evaluate using the protocol of \cite{semoli,waymo}, using an area of whole 100\si{\m}$\times$40\si{\m} \bev~grid around the ego vehicle, considering objects that move faster than 1\si{\m/\s} to be \textit{moving} (difficulty level L2).
CP, TF: network architecture, in the first two lines trained supervised for comparison.
$\dagger$: Results taken from \cite{semoli}.
SF: \lsf\ by SLIM, GF: ground truth \lsf.
}
\label{tab:waymo_full}
\end{table*}

\end{document}